\documentclass{article}

\usepackage{iclr2027_conference,times}

\usepackage{amsmath,amssymb}

\usepackage{amsmath,amsfonts,bm}

\def\eqref#1{equation~\ref{#1}}
\def\1{\bm{1}}

\DeclareMathAlphabet{\mathsfit}{\encodingdefault}{\sfdefault}{m}{sl}
\SetMathAlphabet{\mathsfit}{bold}{\encodingdefault}{\sfdefault}{bx}{n}

\usepackage{xcolor}
\usepackage{colortbl}
\usepackage[hidelinks,hypertexnames=false,hyperfootnotes=false]{hyperref}
\usepackage{url}
\usepackage{booktabs}
\usepackage{tabularx}
\usepackage{multirow}
\usepackage{placeins}
\usepackage{float}
\usepackage{graphicx}

\definecolor{ContourTitleBlue}{HTML}{3F51A3}
\definecolor{ContourTitleGreen}{HTML}{10A89A}
\definecolor{TableRankRed}{HTML}{E45B5B}
\definecolor{TableRankOrange}{HTML}{F09545}
\definecolor{TableRankGold}{HTML}{EABF42}
\definecolor{TableRankYellow}{HTML}{E3C933}

\newcommand{\gradientcontourvla}{%
  \textcolor{ContourTitleBlue!100!ContourTitleGreen}{C}%
  \textcolor{ContourTitleBlue!89!ContourTitleGreen}{o}%
  \textcolor{ContourTitleBlue!78!ContourTitleGreen}{n}%
  \textcolor{ContourTitleBlue!67!ContourTitleGreen}{t}%
  \textcolor{ContourTitleBlue!56!ContourTitleGreen}{o}%
  \textcolor{ContourTitleBlue!44!ContourTitleGreen}{u}%
  \textcolor{ContourTitleBlue!33!ContourTitleGreen}{r}%
  \textcolor{ContourTitleBlue!22!ContourTitleGreen}{V}%
  \textcolor{ContourTitleBlue!11!ContourTitleGreen}{L}%
  \textcolor{ContourTitleBlue!0!ContourTitleGreen}{A}%
}

\DeclareRobustCommand{\contourvla}{%
  \texorpdfstring{\gradientcontourvla}{ContourVLA}%
}

\newcommand{\titlebrand}{%
  \texorpdfstring{%
    \mbox{\raisebox{-0.22em}{\includegraphics[height=1.40em]{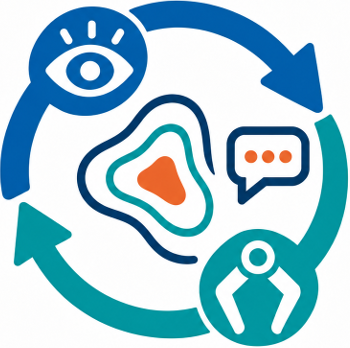}}%
      \hspace{0.42em}%
      {\normalfont\bfseries\gradientcontourvla}}%
  }{ContourVLA}%
}

\title{\raggedright\fontsize{16}{19}\selectfont\titlebrand:\ A Closed-Loop Perception-Action Contour Policy for Generalized Referring Expression Segmentation}
\hypersetup{pdftitle={ContourVLA: A Closed--Loop Perception--Action Contour Policy for Generalized Referring Expression Segmentation}}

\author{%
Ruicheng Zhang\textsuperscript{1,2,*}\hspace{0.75em}
Kaiwen Shen\textsuperscript{1,*}\hspace{0.75em}
Jiaqi Hou\textsuperscript{2,*}\hspace{0.75em}
Shuhan Yang\textsuperscript{2}\hspace{0.75em}
Junchao Huang\textsuperscript{4} \\
\textbf{Kewei Zhang}\textsuperscript{3}\hspace{0.75em}
\textbf{Jun Zhou}\textsuperscript{2}\hspace{0.75em}
\textbf{Li Jiang}\textsuperscript{4}\hspace{0.75em}
\textbf{Shen Zhao}\textsuperscript{1,\(\dagger\)} \\[0.6em]
\normalsize
\textsuperscript{1}Sun Yat-sen University \quad
\textsuperscript{2}Tsinghua University \quad
\textsuperscript{3}Peking University \\
\normalsize
\textsuperscript{4}The Chinese University of Hong Kong, Shenzhen}

\iclrfinalcopy

\begin{document}
\raggedbottom

\maketitle
\lhead{ContourVLA}

\begingroup
\renewcommand{\thefootnote}{}
\footnotetext{\textsuperscript{*}Equal contribution. \quad
\textsuperscript{\(\dagger\)}Corresponding author.}
\endgroup

\vspace{-1.0em}
% BEGIN TEASER: fixed directly before the abstract.
\noindent\includegraphics[width=\linewidth]{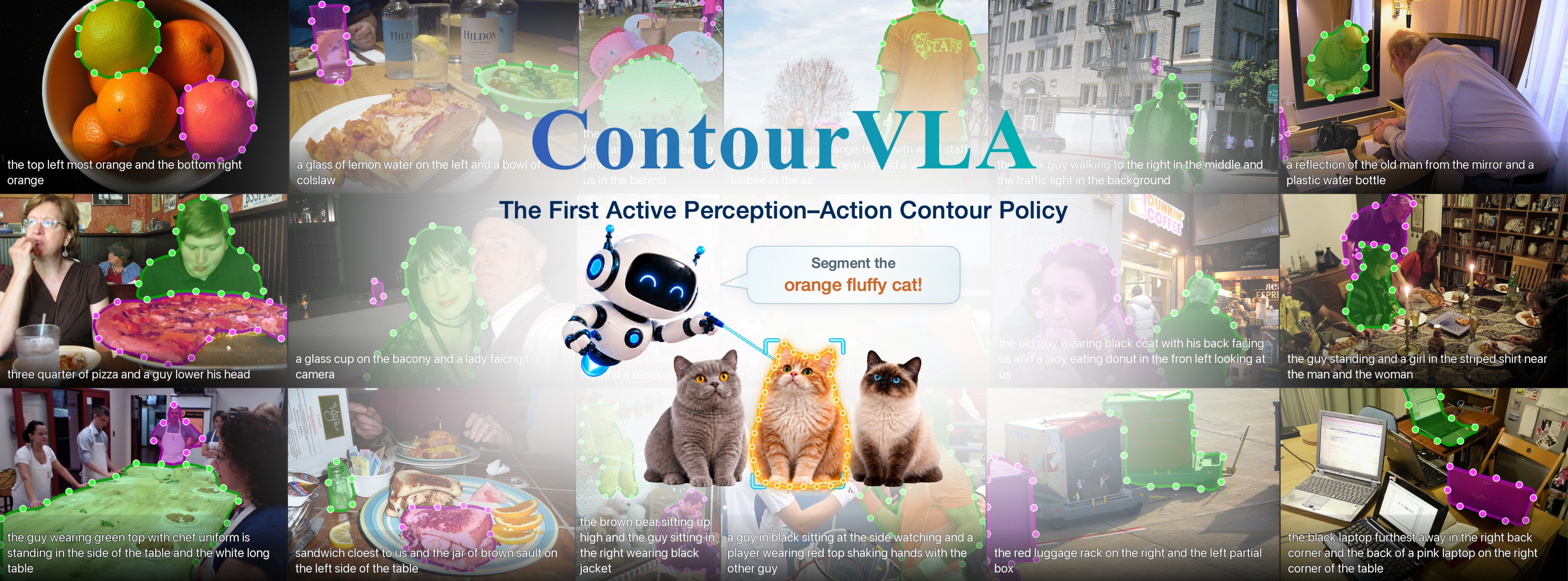}\par
% END TEASER

\begin{abstract}
% \vspace{-0.5em}
Generalized referring expression segmentation (GRES) requires dynamically balancing high-level semantics for identifying a variable number of language-specified referents with fine-grained visual evidence for precise boundary delineation. This requirement challenges existing cascaded vision-language architectures, which typically rely on static feature interfaces and single-pass mask prediction, limiting adaptive perception and geometric correction. We introduce \contourvla, a vision-language-action policy that recasts GRES as a closed-loop visuomotor process, in which editable contours serve as explicit policy states that condition multimodal perception and are updated by geometric action chunks. Evolution-Aware Semantic Scheduling (EASS) couples contour-guided bidirectional boundary sampling with state-conditioned routing of multilevel multimodal features, adapting perception to each contour state. Following supervised initialization, Dustbin-Augmented Entropic Credit Transport GRPO (DECT-GRPO) jointly optimizes discrete grounding and continuous contour actions with instance-level credits. Its rollout rewards and credits are derived from soft prediction-target correspondences that account for false positives and missed targets. \contourvla\ improves gIoU over the strongest evaluated baselines by 8.7, 2.8, and 2.7 points on gRefCOCO val, testA, and testB, respectively, and achieves the highest mIoU across all eight RefCOCO, RefCOCO+, and RefCOCOg splits.
\end{abstract}

% \vspace{-0.5em}
\section{Introduction}
\label{sec:introduction}
% \vspace{-1em}

Generalized referring expression segmentation (GRES) delineates a variable number of objects specified by free-form expressions involving attribute, relational, spatial, and numerical constraints~\citep{liu2023gres,kim2026target}. Recent GRES methods typically adopt a cascaded architecture that connects preselected layers of pretrained vision-language models (VLMs) to segmentation heads through static feature interfaces~\citep{lai2024lisa,xia2024gsva,yan2025deepmoltex,xu2024enhancing}. Beyond the demands of conventional fixed-category segmentation, the GRES segmentation head needs to dynamically balance high-level semantics for identifying language-specified referents and fine-grained visual evidence for recovering precise boundaries. However, static cascaded interfaces limit such adaptive feature acquisition~\citep{luo2025cohd}. Moreover, single-pass mask prediction lacks an updatable intermediate geometric state and an explicit feedback path for revisiting local evidence, hindering the refinement of fine structures and the separation of adjacent instances~\citep{peng2020deepsnake}.

\begin{figure}[t]
    \centering
    \setlength{\abovecaptionskip}{-0.1em}   %调整图片标题与图距离
    \includegraphics[width=0.95\linewidth]{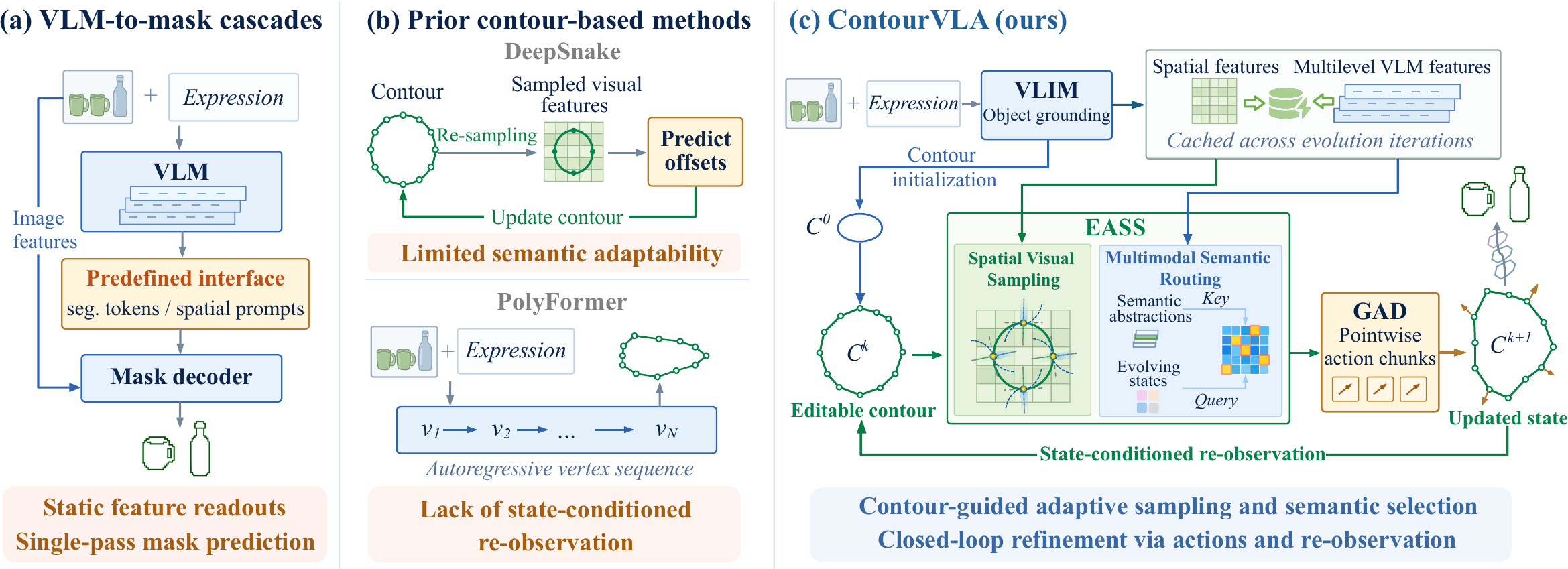}
    \caption{
    \textbf{Paradigm-level comparison of \contourvla\ with prior methods.}
    }
    \label{fig:paradigm}
    \vspace{-1.5em}
\end{figure}

These limitations motivate a shift from static feedforward prediction toward an active vision paradigm, in which the current state guides evidence acquisition and action-induced state updates condition subsequent observations~\citep{bajcsy1988active}. Inspired by this view, we formulate GRES as a text-conditioned visuomotor decision process, where an editable contour serves as an explicit geometric state that regulates multimodal perception and evolves through pointwise geometric actions. Figure~\ref{fig:paradigm} contrasts this formulation with VLM-to-mask cascades and prior contour-based methods.

We introduce \contourvla, a vision-language-action (VLA) policy~\citep{kim2024openvla} that formulates GRES as a closed-loop perception-action process, where multimodal observations are iteratively transformed into geometric actions for contour evolution. \contourvla\ consists of a V-L Interaction Module (VLIM) and a Geometric Action Decoder (GAD). VLIM acts as a perceptual expert that grounds text-specified referents and provides multimodal observations for state-conditioned perception, while GAD serves as an action expert that predicts geometric action chunks from these observations~\citep{zhao2023act}. Executing each chunk updates the contour state and triggers the next perception–action cycle, progressively refining the contour from coarse localization to precise boundaries. \contourvla\ adopts a two-stage training strategy: SFT establishes grounding and contour-trajectory priors, followed by joint GRPO post-training with self-rollout feedback~\citep{shao2024deepseekmath}. This formulation raises two key challenges: adaptive multimodal perception under evolving contour states and effective optimization of the joint policy.

% Geometric refinement requires adaptive integration of spatial and semantic evidence throughout contour evolution. We therefore introduce Evolution-Aware Semantic Scheduling (EASS), which adapts dual-axis multimodal perception through the evolving contour state. On the spatial axis, Spatial Visual Sampling uses contour geometry as a prior to sample bidirectional boundary cues over an adaptive receptive field. On the semantic axis, Multimodal Semantic Routing encodes instance geometry, evolution progress, contour motion, and decoding depth as queries to select the most relevant VLIM representations.

Geometric refinement requires adaptive integration of spatial and semantic evidence throughout contour evolution, a dynamic demand that static feature interfaces cannot accommodate. We therefore introduce Evolution-Aware Semantic Scheduling (EASS), which adapts dual-axis multimodal perception through the evolving contour state. On the spatial axis, Spatial Visual Sampling uses contour
geometry as a prior to sample bidirectional boundary cues over an adaptive receptive field. On the semantic axis, Multimodal Semantic Routing encodes instance geometry, evolution progress, contour motion, and decoding depth as queries to select the most relevant VLIM representations.

Effective joint-policy optimization in GRES requires multi-instance credit attribution under uneven prediction quality, false positives, and missed targets. Standard GRPO instead broadcasts a single rollout-level advantage across all instances, conflating their contributions and preventing targeted
optimization~\citep{shao2024deepseekmath}. We therefore propose Dustbin-Augmented Entropic Credit Transport GRPO (DECT-GRPO), which assigns credit via soft correspondences established by entropic transport, while dustbin entries absorb false-positive predictions and missed targets~\citep{cuturi2013sinkhorn,sarlin2020superglue}. DECT-GRPO preserves module-specific likelihoods for discrete grounding and continuous contour actions, while converting transport-weighted segmentation quality into branchwise centered credit maps assigned to corresponding grounding tokens and contour actions.

Our main contributions are as follows:

\vspace{-0.75em}
\begin{itemize}
% \vspace{-0.25em}
    \item We introduce \contourvla, the first to recast GRES as an active perception-action contour policy in which geometric states bridge multimodal perception and contour actions.
% \vspace{-0.25em}
    \item We propose EASS, a state-conditioned perceptual scheduler that replaces static readouts with spatial visual sampling for adaptive boundary-cue acquisition and multimodal semantic routing for evolution-aware selection of multilevel multimodal features.
% \vspace{-0.25em}
    \item We propose DECT-GRPO, which couples module-specific likelihood optimization with dustbin-augmented entropic transport to derive instance-level credits for grounding tokens and contour actions while accounting for false-positive predictions and missed targets.
\end{itemize}

% \vspace{-0.5em}
\section{Related Work}
\label{sec:related-work}
% \vspace{-1em}

\paragraph{Generalized Referring Expression Segmentation.}
Generalized referring expression segmentation (GRES) aims to identify and
segment a variable number of objects specified by free-form
expressions~\citep{liu2023gres}. Early approaches rely on cross-modal fusion
between separate visual and language
encoders~\citep{yang2022lavt,liu2023gres,luo2025cohd}, while recent methods
typically connect pretrained VLMs with segmentation decoders through
segmentation tokens or spatial
prompts~\citep{lai2024lisa,xia2024gsva,rasheed2024glamm,liu2025segzero,huang2025samr1,lu2026segcompass}.
Despite varied interfaces, these cascades rely on fixed feature readouts and
single-pass prediction, limiting adaptive semantic-spatial evidence selection
and iterative segmentation correction. 
In contrast, \contourvla\ formulates GRES as a closed-loop visuomotor process, where the evolving contour state regulates visual evidence acquisition and semantic abstraction selection, while iterative geometric actions enable
continuous boundary refinement.

% \vspace{-0.75em}
\paragraph{Contour Evolution and Vision-Language-Action Policies.}

Contour-based methods represent boundaries as ordered vertices and refine them through geometric updates. DeepSnake and its
variants~\citep{peng2020deepsnake,gamesnake,MambaContour,zhang2026teams,mambasnake,zhao2023attractive,tang2024progressive} iteratively sample
local visual features for contour refinement, while
PolyFormer~\citep{liu2023polyformer} autoregressively predicts
language-conditioned contour vertices. These methods do not adapt multimodal
evidence to evolving contour states in GRES. Inspired by vision-language-action
models~\citep{kim2024openvla}, \contourvla\ instead recasts contour evolution as a 2D geometric action prediction problem conditioned on vision and language, and solves it through a closed-loop perception-action process.

% \vspace{-0.75em}
\paragraph{Reinforcement Learning for Vision-Language Reasoning and Grounding.}
Reinforcement Learning (RL) has emerged as a potent methodology for enhancing performance on complex visual tasks~\cite{zhang2026kvpo,zhang2025mind,zhang2026robostereo,luo2026learning,guo2026should}.
Seg-Zero~\citep{liu2025segzero} and SAM-R1~\citep{huang2025samr1} apply
GRPO~\citep{shao2024deepseekmath} with localization or mask-quality rewards
to multimodal segmentation and grounding. Standard GRPO, however, shares a
rollout-level advantage across all outputs without explicitly distinguishing
instance quality in the advantage assigned to each prediction.
MCR-GRPO~\citep{han2026mcrgrpo} estimates instance importance through
marginal contribution analysis. DECT-GRPO instead
derives instance-level credit assignments via dustbin-augmented entropy-regularized transport, with dustbin assignments absorbing unmatched predictions and targets.

% \vspace{-0.5em}
\section{Method}
\label{sec:method}
% \vspace{-1em}

\begin{figure}[t]
    \centering
    \setlength{\abovecaptionskip}{-0.5em}   %调整图片标题与图距离
    \includegraphics[width=\linewidth]{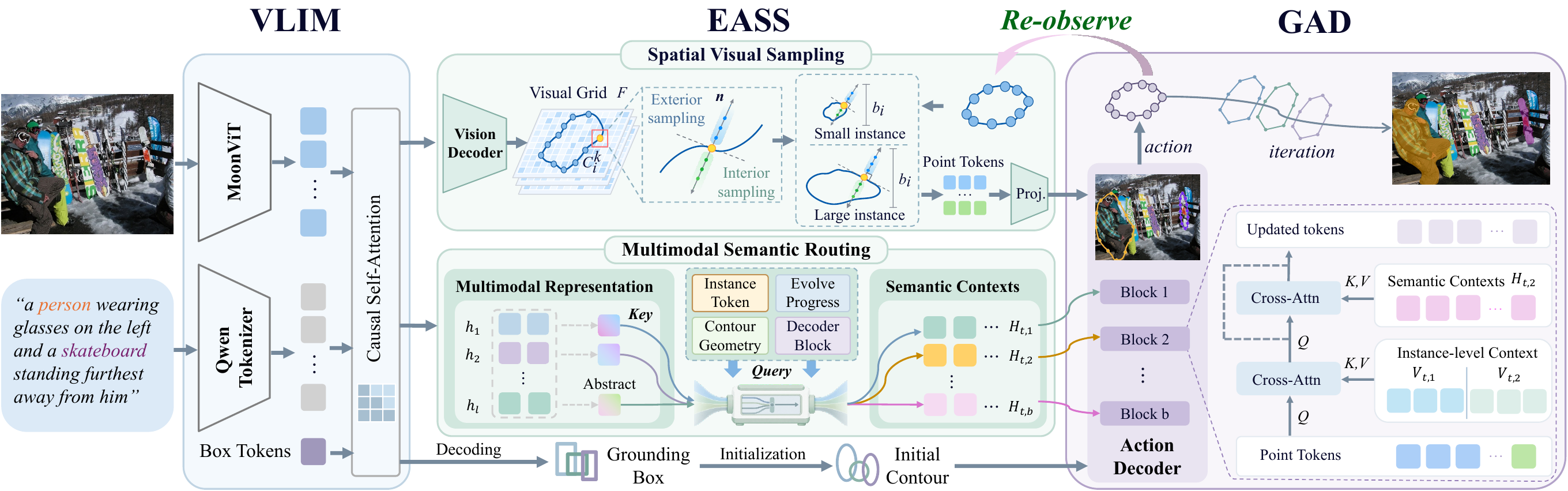}
    \caption{
    \textbf{Overview of the \contourvla\ framework.}
    VLIM predicts grounding boxes for contour initialization and
    provides visual and multilevel multimodal features.
    EASS performs adaptive bidirectional boundary sampling and
    routes semantic representations according to the evolving state.
    GAD combines contour point tokens with instance context and
    routed semantics to predict pointwise geometric action chunks.
    Executing these actions updates the contours and conditions
    subsequent observations, forming a closed perception-action loop.
    The dashed inset illustrates a GAD block.
    }
    \label{fig:framework}
    \vspace{-1em}
\end{figure}

As shown in Figure~\ref{fig:framework}, \contourvla\ formulates GRES as a text-conditioned visuomotor decision process, where editable contours serve as geometric policy states that condition perception and action. Executing action chunks iteratively updates the contour state and closes the perception-action loop (Section~\ref{sec:framework}). 
Within this closed loop, EASS constructs state-conditioned multimodal observations by sampling around the contour and scheduling semantic features (Section~\ref{sec:eass}). GAD uses these observations to predict geometric action chunks that update the contour state (Section~\ref{sec:gad}). Training follows a two-stage strategy: supervised initialization establishes a geometric prior, followed by DECT-GRPO post-training with instance-level credit assignment (Section~\ref{sec:policy-optimization}).

% Formally, at iteration $k$, the process is
% \begin{equation}
% (\underbrace{\mathrm{V}}_{\text{visual input }},
% \underbrace{\mathrm{L}}_{\text{text input }})
% \longrightarrow
% \underbrace{S^k\equiv C^k}_{\text{contour state}}
% \xrightarrow{\pi_{\theta,\phi}}
% \underbrace{\mathrm{A}}_{\text{action chunk }}
% \xrightarrow{\mathcal T}
% \underbrace{S^{k+1}\equiv C^{k+1}}_{\text{updated contour state}}.
% \label{eq:vla-loop}
% \end{equation}

% \vspace{-0.5em}
\subsection{State-Conditioned Perception-Action Loop}
\label{sec:framework}

\vspace{-0.5em}

\textbf{Multimodal perception and contour initialization.}
Given an image-expression pair $(I,x)$, VLIM autoregressively generates a structured grounding sequence $y_{1:L_y}\sim\pi_\theta^y(\cdot\mid I,x)$, which contains reasoning tokens followed by either $M$ box records or a no-target response:
\begin{equation}
y_{1:L_y}=
\begin{cases}
r_{1:L_r}
[\langle\mathrm{box}\rangle
\langle x_1\rangle
\langle y_1\rangle
\langle x_2\rangle
\langle y_2\rangle]_{i=1}^{M}
\langle\mathrm{eos}\rangle, & M>0,\\
r_{1:L_r}\langle\mathrm{no-target}\rangle\langle\mathrm{eos}\rangle, & M=0.
\end{cases}
\label{eq:grounding-format}
\end{equation}
The coordinate tokens in each box record are decoded into a bounding box $b_i$, which initializes an inscribed elliptical contour
$C_i^0=(c_{i,1}^0,\ldots,c_{i,P_i}^0)\in\mathbb R^{P_i\times2}$.
The point budget $P_i$ is determined by the box perimeter to adapt contour resolution to object scale. Meanwhile, VLIM caches a visual feature grid $F$ and multilevel multimodal representations $\{H^\ell\}_{\ell\in\mathcal L}$ for subsequent state-conditioned perception. Region features pooled from $F$ are further contextualized with box geometry to obtain the instance representation $V_i$.

\textbf{Perception-action loop.}
At evolution iteration $k$, EASS constructs state-conditioned observations along spatial and semantic axes. The spatial branch extracts contour-conditioned boundary evidence
$\Phi(F,C_i^k,b_i)$ from the visual grid, while the semantic branch selects block-specific multimodal features $\mathcal U_i^k$ from $\{H^\ell\}_{\ell\in\mathcal L}$. Together with the instance representation $V_i$, these readouts constitute the multimodal observation $O_i^k$ for geometric action chunk prediction $A_i^k$ in GAD:
\begin{equation}
O_i^k=
\left(
\Phi(F,C_i^k,b_i),V_i,\mathcal U_i^k
\right),
\qquad
A_i^k=\pi_\phi(C_i^k,O_i^k),
\qquad
C_i^{k+1}=\mathcal T(C_i^k,A_i^k).
\label{eq:perception-action}
\end{equation}
Here $\mathcal T$ denotes the geometric state transition function. The updated contour $C_i^{k+1}$ initializes the next perception step. Thus, the contour serves as an explicit policy state connecting iterative observation, geometric refinement, and state evolution.

% \vspace{-0.5em}
\subsection{Evolution-Aware Semantic Scheduling}
\label{sec:eass}
\vspace{-0.5em}
The spatial evidence and semantic abstractions required for geometric refinement vary across evolution stages and contour states. We therefore introduce Evolution-Aware Semantic Scheduling (EASS), a dual-axis scheduler that adapts visual evidence acquisition and multimodal feature selection throughout evolution.

% \vspace{-0.5em}
\subsubsection{Spatial Visual Sampling}
\vspace{-0.5em}
At each evolution iteration, the current contour provides a geometric prior for acquiring local visual evidence. Beyond vertex-level feature sampling~\citep{peng2020deepsnake}, EASS dynamically expands the receptive field along each vertex normal by sampling bidirectional interior and exterior features from visual grid $F$. Let $b_i^{\mathrm{size}}=(w_i,h_i)$ denote the size of box $b_i$, and let $n_{i,p}^k$ be the unit normal computed after mapping the contour into box-normalized coordinates. Using nonnegative offsets $\{o_j\}_{j=1}^{J}$, the contour-aligned feature profile is
\begin{equation}
\phi_{i,p}^k=
\operatorname{Concat}_{j=1}^{J}
\left[
F\!\left(c_{i,p}^k-o_j(n_{i,p}^k\odot b_i^{\mathrm{size}})\right),
F\!\left(c_{i,p}^k+o_j(n_{i,p}^k\odot b_i^{\mathrm{size}})\right)
\right].
\label{eq:normal-observation}
\end{equation}
Bidirectional sampling captures complementary interior and exterior boundary
cues, while box-relative scaling adapts the receptive field to object scale. Details are provided in
Appendix~\ref{sec:coordinate-action-details}.

% \vspace{-0.5em}
\subsubsection{Multimodal Semantic Routing}
\vspace{-0.5em}
% Semantic requirements change during contour refinement. Early iterations require stronger object-level semantics to maintain referential consistency, whereas later iterations rely more on fine-grained representations for geometric correction. EASS therefore introduces a key-query router that matches VLIM semantic abstractions to the refinement demands of evolving contour states.

\textbf{Semantic abstractions as keys.} The multilevel multimodal features of VLIM constitute a candidate semantic space. For each layer $\ell$, let $H^\ell$ denote its multimodal token representations. EASS compresses $H^\ell$ into a key that characterizes the semantic abstraction provided by each layer:
\begin{equation}
\kappa_\ell=
\operatorname{Norm}_2
\left(
f_{\mathrm{key}}\!\left(\operatorname{Pool}(H^\ell)+e_\ell\right)
\right),
\label{eq:layer-key}
\end{equation}
where $\operatorname{Norm}_2(z)=z/\|z\|_2$ and $e_\ell$ is a learned layer identity.

\textbf{Evolving states as queries.}
Instance characteristics $g_i$, contour states $c_i^k$, evolution stages $r^k$, and decoder depths $d$ jointly determine the semantic requirements of each refinement step. EASS encodes these factors into queries to select the VLIM layer that provides the most relevant semantic cues.
For object $i$ at evolution iteration $k$ and GAD block $d$, the query is constructed as
\begin{equation}
q_i^{k,d}
=
\operatorname{Norm}_2
\left(
f_q(g_i,c_i^k,r^k,d)
\right).
\label{eq:semantic-query}
\end{equation}
Here $g_i=(b_i,P_i)$ denotes the instance characteristics, where
$b_i=(x_i^1,y_i^1,x_i^2,y_i^2)$ is the normalized grounding box and
$P_i$ is the object-adaptive contour point budget.

\textbf{Evolution-aware key-query routing.}
EASS computes the compatibility between each evolution-conditioned query and VLIM semantic keys, yielding a probability distribution over candidate layers:
\vspace{-0.5em}
\begin{equation}
p_{i,\ell}^{k,d}
=
\operatorname{Softmax}_{\ell\in\mathcal L}
\left(
\langle q_i^{k,d},\kappa_\ell\rangle
+\alpha_u B_{k,d,\ell}
\right),
\label{eq:state-routing}
\end{equation}
Here $p_{i,\ell}^{k,d}$ denotes the layer-selection probability for VLIM layer $\ell$ of object $i$ at evolution iteration $k$ and GAD block $d$. To stabilize early routing, the prior $B$ initializes an ordered correspondence between evolution-depth slots and VLIM layers, while its weight gradually decays to allow contour-conditioned query-key compatibility to dominate.
To retain discrete layer selection in the forward pass while enabling differentiable router optimization, EASS applies hard routing with a straight-through estimator:
\begin{equation}
\widetilde p_i^{k,d}
=
\operatorname{sg}\bigl(
\operatorname{onehot}(\arg\max_\ell p_{i,\ell}^{k,d})
-p_i^{k,d}
\bigr)
+p_i^{k,d},
\qquad
U_i^{k,d}
=
\operatorname{Proj}
\bigl(
\sum_{\ell\in\mathcal L}
\widetilde p_{i,\ell}^{k,d}H^\ell
\bigr),
\label{eq:routed-feature}
\end{equation}
where $\operatorname{sg}$ denotes stop-gradient and $U_i^{k,d}$ is the routed multimodal feature delivered to the $d$-th GAD block. $\operatorname{Proj}$ projects the selected VLIM representation into the corresponding GAD feature space. Recomputing the query after each contour update adapts the routed semantic abstraction throughout multistep contour evolution.
Further details are provided in Appendix~\ref{sec:implementation-details}.

% \vspace{-0.5em}
\subsection{Geometric Action Chunking}
\label{sec:gad}
\vspace{-0.5em}

GAD decodes EASS observations into continuous pointwise displacement actions. Each contour vertex is represented as a geometric token based on the contour-conditioned visual evidence $\phi_{i,p}^k$ from Spatial Visual Sampling and contour geometry:
\begin{equation}
Z_{i,p}^{k,0}
=
\operatorname{Proj}
\left(
\operatorname{Concat}
[
\phi_{i,p}^{k},
c_{i,p}^{k},
\operatorname{PE}_{\mathrm{box}}(c_{i,p}^{k};b_i),
\operatorname{PE}_{\mathrm{cyc}}(p/P_i)
]
\right).
\label{eq:point-token}
\end{equation}
where $\operatorname{PE}_{\mathrm{box}}(c_{i,p}^{k};b_i)$ encodes box-relative vertex coordinates, and $\operatorname{PE}_{\mathrm{cyc}}(p/P_i)$ preserves the cyclic positions of vertices along the closed contour. Stacking the $P_i$ point tokens yields the contour representation $Z_i^{k,0}=[Z_{i,1}^{k,0},\ldots,Z_{i,P_i}^{k,0}]^\top\in\mathbb R^{P_i\times h}$, which is further refined with instance-level context $V_i$. At each GAD block, contour tokens interact with the semantic representation $U_i^{k,d}$:
\begin{equation}
\overline Z_i^{k,d}
=
\operatorname{CrossAttn}_d(Z_i^{k,d-1},V_i),
\qquad
Z_i^{k,d}
=
\overline Z_i^{k,d}
+
\operatorname{CrossAttn}_d
(\overline Z_i^{k,d},U_i^{k,d}).
\label{eq:semantic-injection}
\end{equation}
The action decoder then predicts two-dimensional displacement means for each vertex in parallel~\citep{zhao2023act}, which updates the contour state and triggers the next iteration.

% \vspace{-0.5em}
\subsection{Policy Optimization with DECT-GRPO}
\label{sec:policy-optimization}
\vspace{-0.5em}

\begin{figure}[t]
\centering
\setlength{\abovecaptionskip}{-0.50em}   %调整图片标题与图距离
\includegraphics[width=\linewidth]{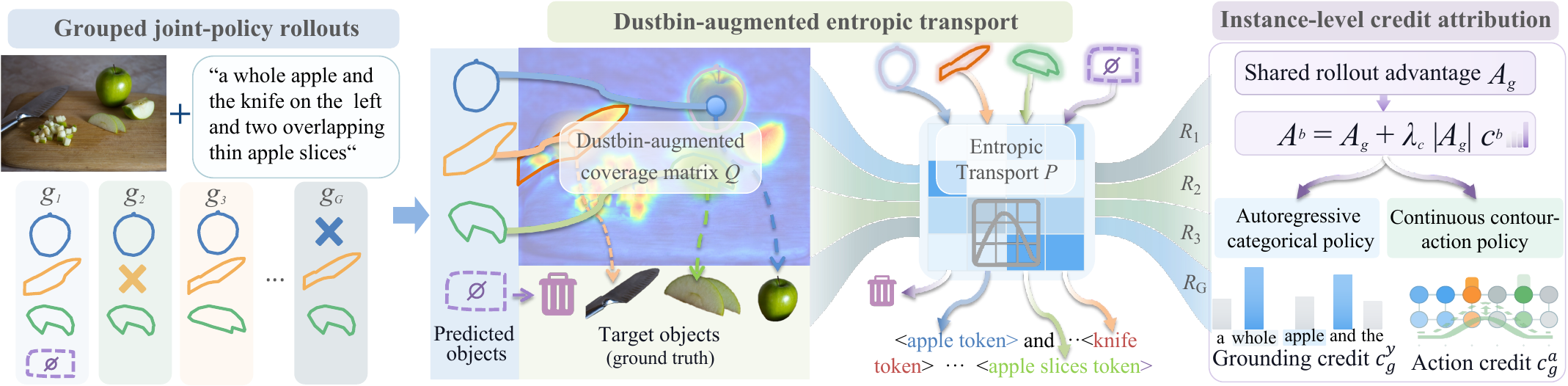}
\caption{\textbf{Overview of DECT-GRPO.} Dustbin-augmented entropic transport establishes soft
prediction–target correspondences and assigns instance-level credits to grounding tokens and contour actions for joint policy optimization.}
\label{fig:dect_grpo}
\vspace{-1em}
\end{figure}

We initialize VLIM with teacher-forced grounding and GAD with positional supervision along reference contour trajectories, regularized by auxiliary geometric and routing objectives (Appendix~\ref{sec:sft-losses}). 
Building on this initialization, DECT-GRPO (Fig.~\ref{fig:dect_grpo}) leverages dustbin-augmented entropic transport to derive instance-level credit attribution and transport-based rewards for jointly optimizing grounding and geometric actions.

\textbf{Grouped joint-policy rollouts.}
For each input $(I,x)$, VLIM samples $G$ grounding outputs that define the
initial contours, after which GAD generates continuous geometric action
chunks. The grounding branch follows an autoregressive categorical policy,
while the action branch follows a bounded continuous policy based on a
$\tanh$-transformed Gaussian. Their likelihood definitions and
rollout-replay details are provided in
Appendix~\ref{sec:joint-rl-replay}.

\paragraph{Dustbin-augmented entropic transport.}
Instances within a GRES rollout may differ substantially in segmentation quality, yet standard GRPO assigns all outputs the same rollout-level advantage without explicitly distinguishing instance quality in the assigned advantage~\citep{shao2024deepseekmath}. DECT-GRPO therefore uses dustbin-augmented entropic transport to establish soft prediction-target correspondences for reward construction and instance-level credit attribution, with virtual null targets and predictions absorbing false positives and missed targets.

For rollout $g$ among the $G$ samples of an input, let $M_g$ and $N$ denote the predicted and target instance counts, respectively. With $Q_{g,ij}$ denoting the mask IoU between prediction $i$ and target $j$, the pairwise coverage matrix $Q_g=[Q_{g,ij}]\in\mathbb R^{M_g\times N}$ is augmented as
\begin{equation}
\overline Q_g=
\begin{bmatrix}
Q_g & b_\varnothing\mathbf 1_{M_g}\\
b_\varnothing\mathbf 1_N^\top & b_\varnothing
\end{bmatrix}
\in\mathbb R^{(M_g+1)\times(N+1)},
\label{eq:dect-augmentation}
\end{equation}
where $b_\varnothing$ is the null-match score. The last column absorbs unmatched predictions, while the last row accounts for missed targets.
Let $P\in\mathbb R^{(M_g+1)\times(N+1)}$ denote the soft transport plan, where $P_{ij}$ is the matching mass assigned between prediction $i$ and target $j$, including dustbin matches. DECT-GRPO obtains the globally consistent soft assignment through entropy-regularized optimal transport:
\begin{equation}
P_g^\star=
\arg\max_{P\in\mathcal U_g}
\biggl[
\sum_{i,j}P_{ij}\overline Q_{g,ij}
-\varepsilon\sum_{i,j}P_{ij}\log P_{ij}
\biggr].
\label{eq:dect-transport}
\end{equation}
Here $\mathcal U_g$ is the set of nonnegative transport plans satisfying the prescribed prediction and target marginals. The entropy term prevents ambiguous overlaps from collapsing prematurely to hard correspondences, preserving uncertainty among plausible matches and yielding smoother instance-level credit. The marginal constraints enforce global consistency across predictions and targets, so $P_{g,ij}^\star$ represents a globally normalized soft correspondence rather than an independent pairwise score. Mass assigned to the null target denotes unmatched predictions and lowers instance credit, whereas mass from the null prediction denotes missed targets and reduces target coverage. Its Sinkhorn solution is derived in Appendix~\ref{sec:dect-transport-derivation}.

\paragraph{Transport-based reward.}

A reliable rollout reward should jointly capture contour accuracy, target coverage, and prediction-target correspondence. DECT-GRPO therefore evaluates segmentation quality from complementary prediction and target perspectives under the soft transport plan $P_g^\star$:
\begin{equation}
\resizebox{0.94\linewidth}{!}{$\displaystyle
\mathrm{Prec}_g=\frac{1}{M_g}\sum_{i=1}^{M_g}\frac{\sum_{j=1}^{N}P^\star_{g,ij}Q_{g,ij}}{\sum_{j=1}^{N+1}P^\star_{g,ij}},\quad
\mathrm{Rec}_g=\frac{1}{N}\sum_{j=1}^{N}\frac{\sum_{i=1}^{M_g}P^\star_{g,ij}Q_{g,ij}}{\sum_{i=1}^{M_g+1}P^\star_{g,ij}},\quad
R_g=\frac{2\,\mathrm{Prec}_g\,\mathrm{Rec}_g}{\mathrm{Prec}_g+\mathrm{Rec}_g}
$}
\label{eq:transport-reward}
\end{equation}
Here, $\mathrm{Prec}_g$ and $\mathrm{Rec}_g$ denote transport-aware precision
and recall under $P_g^\star$. We set $R_g=0$ when
$\mathrm{Prec}_g+\mathrm{Rec}_g=0$; empty-set cases are handled separately in
Appendix~\ref{sec:rl-protocol}. Soft transport stabilizes the reward under
ambiguous correspondences by avoiding discontinuities from hard matching.
Dustbin mass remains in the normalization but receives no overlap credit;
therefore, unmatched predictions reduce $\mathrm{Prec}_g$, while missed
targets reduce $\mathrm{Rec}_g$. The resulting harmonic mean jointly captures contour quality, correspondence uncertainty, and target coverage.

\paragraph{Instance-level advantage modulation.}

Predictions within a rollout may differ in segmentation quality, whereas a shared rollout advantage applies the same scalar feedback to all outputs. DECT-GRPO uses transport-derived instance credits to modulate this shared advantage at the corresponding grounding-token spans and contour-action positions, introducing instance-dependent feedback while retaining the rollout-level signal.

For each predicted instance $i$, DECT-GRPO computes transport-weighted credit and broadcasts its normalized value to the corresponding grounding-token span $S_{g,i}$ and contour-action indices $\mathcal D_{g,i}$. Branch-wise centering and scaling produce the credit maps:
\begin{equation}
e_{g,i}=
\frac{\sum_{j=1}^{N}P^\star_{g,ij}Q_{g,ij}}
{\sum_{j=1}^{N+1}P^\star_{g,ij}},\qquad
\mathbf c_g^y[S_{g,i}]=\operatorname{Norm}_y(e_{g,i}),\qquad
\mathbf c_g^a[\mathcal D_{g,i}]=\operatorname{Norm}_a(e_{g,i}).
\label{eq:credit-map}
\end{equation}
The normalization is performed over branch outputs, yielding zero-centered credit maps that preserve the branch-wise mean advantage. The resulting output-specific advantages are
\begin{equation}
A^b_{g,\nu}
=
A_g+\lambda_c|A_g|\,\mathbf c_g^b[\nu],
\qquad b\in\{y,a\},
\label{eq:instance-advantage}
\end{equation}
Answer tokens outside instance spans retain the shared rollout advantage $A_g$.
The output-count-weighted normalization preserves the branch-wise mean
advantage. Since $\mathbf c_g^b[\nu]\in[-1,1]$ and we use
$\lambda_c=0.5$, the instance-modulated advantages retain the sign of
$A_g$. Higher-credit instances receive larger positive advantages when
$A_g>0$ and smaller-magnitude negative advantages when $A_g<0$. Thus,
DECT-GRPO provides bounded, mean-preserving modulation of shared rollout
feedback rather than independent instance-level reward or penalty decisions.

\paragraph{Joint policy optimization.}

Let $\rho^y_{g,n}$ and $\rho^a_{g,\xi}$ denote the current-to-behavior
likelihood ratios for grounding tokens and contour actions, respectively.
The joint objective is
\begin{equation}
\resizebox{0.94\linewidth}{!}{$\displaystyle
\mathcal L_{\mathrm{DECT}}=\frac{1}{G}\sum_{g=1}^{G}\left[
-\frac{1}{L_g}\sum_{n=1}^{L_g}\mathcal S_{\mathrm{clip}}(\rho^y_{g,n},A^y_{g,n})
-\frac{\lambda_a}{|\mathcal D_g|}\sum_{\xi\in\mathcal D_g}\mathcal S_{\mathrm{clip}}(\rho^a_{g,\xi},A^a_{g,\xi})
+\beta_{\mathrm{KL}}\mathcal K_g\right]
$}
\label{eq:joint-loss}
\end{equation}
Here $\lambda_a$ balances the two branches, and $\mathcal K_g$ regularizes
the grounding policy toward the reference model. Likelihood ratios, the
clipped surrogate, and replay details are given in
Appendix~\ref{sec:rl-protocol}.

% \vspace{-0.5em}
\section{Experiments}
\label{sec:experiments}
% \vspace{-0.5em}
\subsection{Experimental Setup}
\label{sec:experimental-setup}
% \vspace{-0.5em}
\textbf{Datasets and metrics.}
We evaluate GRES on gRefCOCO~\citep{liu2023gres} and conventional referring expression segmentation (RES) on RefCOCO, RefCOCO+~\citep{yu2016modeling}, and RefCOCOg~\citep{mao2016generation}. We report gIoU and cIoU for GRES, and mIoU and cIoU for RES.

\textbf{Training setup.}
\contourvla\ initializes its visual and language backbones from MoonViT and Qwen2.5-3B-Instruct, respectively. Grounding training on LocateAnything-Data~\citep{wang2026locateanything} is followed by SFT and DECT-GRPO jointly on the training sets of gRefCOCO, RefCOCO, RefCOCO+, and RefCOCOg. A single resulting checkpoint is evaluated across all benchmark splits. Training and evaluation details are provided in Appendix~\ref{app:training_eval}.

% \vspace{-0.5em}
\subsection{Comparison with Existing Methods}
\label{sec:comparison}
\vspace{-0.5em}

Table~\ref{tab:main_comparison} compares \contourvla\ with existing methods on
GRES and conventional RES benchmarks under their respective model and training
configurations. All results are obtained from our own evaluation runs, while
component-level contributions are examined through the ablations in
Section~\ref{sec:ablation}.

\textbf{GRES results.}
On gRefCOCO, \contourvla\ achieves 83.5, 80.3, and 74.4 gIoU on val,
testA, and testB, respectively, exceeding the strongest evaluated baseline on
each split by 8.7, 2.8, and 2.7 points. It also achieves the highest cIoU on
val and testA, while its testB cIoU of 71.8 is just 0.1 point below HiMTok.
These results combine consistent gains in expression-averaged quality with
competitive cumulative pixel overlap, rather than showing an advantage under
only one aggregation criterion. Since gIoU also accounts for no-target
expressions, Section~\ref{sec:ablation} further examines no-target accuracy and
instance F1 to characterize rejection and instance-level segmentation more
directly.

\textbf{RES results.}
On conventional RES benchmarks, the same jointly trained checkpoint achieves
the highest mIoU on all eight evaluated splits. Improvements over the strongest
baselines range from 1.3 to 1.8 points on RefCOCO and from 2.8 to 3.2 points on
RefCOCO+, with additional gains of 0.7 and 0.3 points on RefCOCOg val and test,
respectively. These results extend the evidence beyond the generalized setting:
the shared contour policy also performs well on conventional referring
segmentation. Importantly, all benchmarks use the same checkpoint without
further benchmark-specific fine-tuning, supporting the applicability of one
policy to both task formulations under the joint-training protocol.

\begin{table}[htb]
\centering
\caption{Comparison with methods across GRES and RES benchmarks (\%). Red, orange, gold, and yellow shading denote the first- through fourth-best scores in each column, respectively.}
\label{tab:main_comparison}
\setlength{\tabcolsep}{1.5pt}
\renewcommand{\arraystretch}{1.10}
\resizebox{\linewidth}{!}{%
\begin{tabular}{l*{22}{c}}
\toprule
\multirow{3}{*}{Method} & \multicolumn{6}{c}{gRefCOCO} & \multicolumn{6}{c}{RefCOCO} & \multicolumn{6}{c}{RefCOCO+} & \multicolumn{4}{c}{RefCOCOg} \\
\cmidrule(lr){2-7} \cmidrule(lr){8-13} \cmidrule(lr){14-19} \cmidrule(lr){20-23}
 & \multicolumn{2}{c}{val} & \multicolumn{2}{c}{testA} & \multicolumn{2}{c}{testB} & \multicolumn{2}{c}{val} & \multicolumn{2}{c}{testA} & \multicolumn{2}{c}{testB} & \multicolumn{2}{c}{val} & \multicolumn{2}{c}{testA} & \multicolumn{2}{c}{testB} & \multicolumn{2}{c}{val} & \multicolumn{2}{c}{test} \\
\cmidrule(lr){2-3} \cmidrule(lr){4-5} \cmidrule(lr){6-7} \cmidrule(lr){8-9} \cmidrule(lr){10-11} \cmidrule(lr){12-13} \cmidrule(lr){14-15} \cmidrule(lr){16-17} \cmidrule(lr){18-19} \cmidrule(lr){20-21} \cmidrule(lr){22-23}
 & gIoU & cIoU & gIoU & cIoU & gIoU & cIoU & mIoU & cIoU & mIoU & cIoU & mIoU & cIoU & mIoU & cIoU & mIoU & cIoU & mIoU & cIoU & mIoU & cIoU & mIoU & cIoU \\
\midrule
Grounded-SAM~\citep{ren2024groundedsam} & 46.0 & 44.1 & 50.8 & 48.7 & 43.5 & 41.3 & 46.2 & 44.4 & 48.6 & 47.0 & 42.9 & 40.9 & 38.2 & 36.5 & 42.4 & 40.5 & 33.0 & 30.8 & 44.1 & 42.0 & 44.5 & 43.0 \\
LISA-7B~\citep{lai2024lisa} & 61.4 & 61.7 & 66.0 & 68.5 & 58.8 & 60.6 & 68.1 & 74.6 & 70.6 & 79.3 & 64.5 & 72.3 & 54.8 & 65.1 & 60.3 & 70.8 & 48.4 & 58.1 & 63.1 & 67.9 & 66.2 & 70.6 \\
LISA-13B~\citep{lai2024lisa} & 63.4 & 62.9 & 68.1 & 69.6 & 61.8 & 62.2 & 75.3 & 76.0 & 77.8 & 78.9 & 71.7 & 72.9 & 64.9 & 65.0 & 70.1 & 70.1 & 58.9 & 58.1 & 65.0 & 69.6 & 66.0 & 70.4 \\
PerceptionGPT-7B~\citep{pi2024perceptiongpt} & 61.5 & 59.9 & 66.1 & 64.3 & 58.9 & 56.8 & 70.7 & 74.7 & 73.9 & 78.7 & 66.8 & 71.7 & 64.0 & 68.1 & 69.4 & 73.8 & 56.8 & 61.0 & 65.5 & 70.2 & 67.2 & 71.5 \\
PerceptionGPT-13B~\citep{pi2024perceptiongpt} & 62.2 & 60.4 & 66.9 & 64.6 & 59.0 & 57.4 & 70.8 & 75.7 & 74.5 & 79.4 & 67.9 & 72.0 & 64.4 & 69.2 & 69.8 & 74.0 & 57.6 & 62.0 & 66.4 & 71.1 & 67.4 & 72.0 \\
LaSagnA-7B~\citep{wei2024lasagna} & 32.4 & 38.1 & 47.3 & 50.4 & 38.9 & 42.1 & 72.8 & 76.8 & 74.2 & 78.8 & 70.7 & 73.8 & 61.4 & 66.4 & 65.3 & 70.6 & 57.3 & 60.1 & 66.1 & 70.5 & 67.3 & 72.0 \\
OMG-LLaVA-7B~\citep{zhang2024omgllava} & 36.1 & 39.3 & 50.1 & 52.4 & 42.2 & 43.7 & 73.5 & 77.8 & 75.7 & 80.3 & 69.3 & 74.1 & 64.6 & 69.3 & 68.6 & 73.0 & 58.8 & 63.0 & 68.1 & 73.0 & 68.2 & 73.2 \\
GroundHog-7B~\citep{zhang2024groundhog} & 66.7 & 64.8 & 71.4 & 71.4 & 64.0 & 62.7 & 74.0 & 78.3 & 75.4 & 80.1 & 71.6 & 75.7 & 66.2 & 70.2 & 70.6 & 75.0 & 60.6 & 64.8 & 69.6 & 74.2 & 69.8 & 74.6 \\
GLaMM-7B~\citep{rasheed2024glamm} & 68.7 & 66.9 & 73.5 & 71.6 & 66.2 & 64.1 & 75.0 & 79.4 & 78.8 & \cellcolor{TableRankOrange!30}83.5 & 72.7 & 76.9 & 67.8 & 72.7 & 74.1 & 78.7 & 60.2 & 64.5 & 69.8 & 74.6 & 70.6 & 74.7 \\
UniRES-7B~\citep{wang2024unires} & \cellcolor{TableRankYellow!15}74.4 & \cellcolor{TableRankYellow!15}69.9 & \cellcolor{TableRankGold!22}76.0 & \cellcolor{TableRankYellow!15}74.5 & \cellcolor{TableRankGold!22}69.8 & 66.6 & 75.8 & 80.1 & 77.2 & 82.0 & 71.7 & 75.8 & 66.9 & 71.8 & 71.8 & 76.0 & 60.0 & 64.4 & 69.0 & 73.7 & 69.5 & 74.1 \\
GSVA-7B~\citep{xia2024gsva} & 66.5 & 63.3 & 71.1 & 69.9 & 61.9 & 60.5 & 72.1 & 76.7 & 74.3 & 79.6 & 68.0 & 73.0 & 64.0 & 68.2 & 69.1 & 73.1 & 58.5 & 62.3 & 61.9 & 66.6 & 62.4 & 67.5 \\
GSVA-13B~\citep{xia2024gsva} & 67.6 & 64.0 & 71.8 & 70.5 & 63.8 & 61.3 & 73.5 & 77.9 & 75.7 & 80.6 & 69.9 & 74.6 & 65.7 & 69.9 & 70.9 & 74.9 & 59.7 & 63.6 & 63.2 & 68.2 & 63.5 & 69.0 \\
SAM4MLLM-7B~\citep{chen2024sam4mllm} & 68.9 & 67.0 & 73.6 & 71.8 & 66.3 & 64.2 & 75.4 & 79.7 & 78.1 & 82.8 & 71.8 & 75.9 & 69.1 & 73.3 & 73.4 & 77.8 & 61.1 & 66.0 & 70.0 & 74.8 & 71.1 & 75.7 \\
SegLLM-7B~\citep{wang2025segllm} & 69.8 & 67.9 & 74.6 & 72.5 & 67.4 & 65.2 & 75.7 & 80.0 & 77.0 & 81.6 & 71.0 & 75.1 & 66.1 & 70.1 & 68.5 & 72.9 & 57.9 & 62.6 & 68.0 & 72.8 & 69.3 & 73.5 \\
UniPixel-7B~\citep{liu2025unipixel} & 70.6 & 68.8 & 75.3 & 73.4 & 68.0 & 65.9 & 76.3 & 80.7 & 78.5 & \cellcolor{TableRankGold!22}83.1 & 73.0 & 77.3 & 70.7 & 75.5 & 75.6 & 78.4 & 65.5 & 69.7 & 72.1 & 76.2 & 72.4 & 77.1 \\
UFO-8B~\citep{tang2025ufo} & 71.0 & 69.1 & \cellcolor{TableRankYellow!15}75.6 & 73.8 & 68.4 & 66.3 & 76.5 & \cellcolor{TableRankYellow!15}81.1 & 77.9 & 82.6 & 73.7 & \cellcolor{TableRankYellow!15}78.6 & 72.8 & \cellcolor{TableRankOrange!30}77.1 & 75.9 & \cellcolor{TableRankYellow!15}80.3 & \cellcolor{TableRankYellow!15}67.9 & \cellcolor{TableRankRed!38}\textbf{72.7} & 72.6 & \cellcolor{TableRankYellow!15}76.7 & 73.1 & \cellcolor{TableRankYellow!15}77.3 \\
LIRA-8B~\citep{li2025lira} & 36.7 & 40.9 & 50.4 & 52.4 & 42.4 & 44.9 & 77.3 & \cellcolor{TableRankOrange!30}81.4 & 78.9 & 82.4 & 73.5 & 77.9 & 71.8 & \cellcolor{TableRankYellow!15}76.1 & 76.6 & \cellcolor{TableRankOrange!30}80.8 & 66.0 & \cellcolor{TableRankGold!22}70.6 & 73.8 & \cellcolor{TableRankOrange!30}78.5 & 73.5 & \cellcolor{TableRankOrange!30}78.3 \\
HiMTok-8B~\citep{wang2025himtok} & 72.1 & \cellcolor{TableRankGold!22}70.4 & 73.4 & \cellcolor{TableRankGold!22}74.9 & \cellcolor{TableRankOrange!30}71.7 & \cellcolor{TableRankRed!38}\textbf{71.9} & \cellcolor{TableRankGold!22}80.0 & \cellcolor{TableRankRed!38}\textbf{85.9} & \cellcolor{TableRankGold!22}80.7 & 81.7 & \cellcolor{TableRankGold!22}78.6 & \cellcolor{TableRankRed!38}\textbf{83.9} & \cellcolor{TableRankGold!22}74.1 & \cellcolor{TableRankRed!38}\textbf{80.5} & \cellcolor{TableRankYellow!15}76.8 & \cellcolor{TableRankRed!38}\textbf{83.7} & \cellcolor{TableRankGold!22}70.0 & \cellcolor{TableRankOrange!30}72.4 & \cellcolor{TableRankOrange!30}75.9 & \cellcolor{TableRankRed!38}\textbf{80.1} & \cellcolor{TableRankGold!22}75.9 & \cellcolor{TableRankRed!38}\textbf{78.9} \\
RAS-13B~\citep{cao2025ras} & \cellcolor{TableRankGold!22}74.6 & \cellcolor{TableRankOrange!30}70.5 & \cellcolor{TableRankOrange!30}77.5 & \cellcolor{TableRankOrange!30}77.0 & \cellcolor{TableRankYellow!15}69.4 & \cellcolor{TableRankGold!22}67.8 & 76.5 & \cellcolor{TableRankYellow!15}81.1 & 79.1 & \cellcolor{TableRankRed!38}\textbf{83.8} & 74.5 & \cellcolor{TableRankGold!22}78.8 & 70.4 & 75.4 & 75.5 & 79.7 & 65.7 & \cellcolor{TableRankYellow!15}70.5 & 71.1 & 76.0 & 73.2 & \cellcolor{TableRankGold!22}77.6 \\
Seg-R1-7B~\citep{you2025segr1} & 60.6 & 58.5 & 65.3 & 63.1 & 57.7 & 55.8 & 69.9 & 74.3 & 74.2 & 78.8 & 63.3 & 67.5 & 58.1 & 63.0 & 66.3 & 71.0 & 53.1 & 57.9 & 66.9 & 71.2 & 67.1 & 71.2 \\
Seg-Zero-7B~\citep{liu2025segzero} & 16.5 & 26.3 & 44.7 & 45.7 & 40.2 & 41.3 & \cellcolor{TableRankYellow!15}78.7 & 76.4 & \cellcolor{TableRankGold!22}80.7 & 80.3 & \cellcolor{TableRankYellow!15}75.6 & 72.5 & \cellcolor{TableRankYellow!15}73.1 & 69.4 & \cellcolor{TableRankGold!22}77.2 & 76.2 & 67.1 & 62.7 & \cellcolor{TableRankYellow!15}75.1 & 72.9 & \cellcolor{TableRankYellow!15}75.5 & 72.6 \\
VisionReasoner-7B~\citep{liu2025visionreasoner} & 27.0 & 39.2 & 58.3 & 65.1 & 48.6 & 51.8 & 76.7 & 73.4 & 78.9 & 77.6 & 72.3 & 67.6 & 69.6 & 63.8 & 74.9 & 70.9 & 61.9 & 55.5 & 71.6 & 67.7 & 71.3 & 68.4 \\
SAM-R1-7B~\citep{huang2025samr1} & 68.5 & 66.4 & 73.1 & 71.2 & 65.6 & 63.8 & 72.4 & 76.3 & 74.8 & 79.5 & 68.7 & 72.9 & 65.0 & 69.8 & 70.2 & 74.8 & 59.3 & 63.4 & \cellcolor{TableRankYellow!15}75.1 & 72.2 & 75.4 & 73.3 \\
Text4Seg-7B~\citep{lan2025text4seg} & 73.6 & 67.9 & 74.1 & 72.8 & 66.0 & 64.8 & 74.5 & 79.1 & 77.3 & 82.2 & 71.9 & 76.0 & 67.6 & 72.3 & 73.1 & 77.3 & 61.7 & 66.1 & 67.3 & 72.1 & 69.4 & 73.7 \\
Text4Seg-13B~\citep{lan2025text4seg} & \cellcolor{TableRankOrange!30}74.8 & 69.8 & 75.1 & 74.3 & 68.0 & \cellcolor{TableRankYellow!15}67.1 & 75.7 & 80.1 & 78.0 & 82.7 & 72.9 & 77.1 & 69.5 & 73.5 & 73.8 & 78.6 & 63.3 & 67.6 & 69.5 & 74.1 & 70.7 & 74.8 \\
RESAnything-12B~\citep{wang2025resanything} & 54.8 & 52.6 & 48.3 & 46.2 & 48.3 & 46.4 & 64.0 & 68.3 & 67.5 & 72.2 & 65.9 & 70.0 & 55.9 & 60.7 & 61.3 & 65.7 & 47.6 & 52.2 & 55.6 & 59.8 & 56.1 & 61.0 \\
SAM3 (top-1 / union)~\citep{carion2025sam3} & 68.1 & 24.0 & 39.9 & 23.1 & 49.2 & 36.4 & 61.0 & 54.8 & 62.5 & 59.1 & 59.8 & 51.9 & 48.5 & 42.9 & 53.8 & 50.0 & 43.6 & 36.5 & 62.5 & 55.5 & 61.7 & 54.4 \\
RefAM-SD~\citep{kukleva2026refam} & 27.0 & 24.8 & 31.6 & 29.7 & 24.4 & 22.0 & 50.1 & 41.9 & 52.5 & 44.8 & 46.3 & 38.4 & 40.4 & 33.5 & 46.1 & 38.0 & 35.0 & 28.9 & 45.1 & 37.3 & 46.1 & 39.3 \\
RefAM-Flux-12B~\citep{kukleva2026refam} & 27.6 & 25.5 & 32.4 & 30.2 & 24.8 & 22.9 & 52.2 & 42.7 & 55.6 & 46.2 & 49.6 & 40.8 & 43.1 & 34.4 & 48.6 & 38.7 & 37.6 & 30.5 & 48.1 & 39.6 & 48.6 & 41.1 \\
LGD-7B~\citep{li2026lgd} & 24.0 & 21.9 & 28.7 & 26.8 & 21.4 & 19.2 & 45.4 & 34.5 & 48.8 & 37.8 & 39.3 & 30.8 & 47.2 & 36.7 & 54.6 & 44.4 & 38.1 & 30.1 & 47.3 & 36.6 & 48.1 & 37.7 \\
LGD+DINO-7B~\citep{li2026lgd} & 28.5 & 26.6 & 33.2 & 31.0 & 25.8 & 23.7 & 49.5 & 42.7 & 54.7 & 48.2 & 41.0 & 36.3 & 49.6 & 47.3 & 58.4 & 50.9 & 38.7 & 34.4 & 50.3 & 45.5 & 51.1 & 47.7 \\
RefChess~\citep{tong2026refchess} & 29.3 & 27.4 & 33.9 & 32.1 & 26.7 & 24.6 & 55.2 & 48.5 & 57.9 & 52.5 & 50.0 & 43.6 & 48.6 & 41.2 & 54.6 & 48.3 & 39.6 & 32.7 & 50.6 & 42.9 & 51.3 & 44.2 \\
Early Semantic Grounding-9B~\citep{he2026early} & 30.1 & 28.0 & 34.5 & 32.8 & 27.2 & 25.3 & 54.0 & 50.2 & 56.1 & 53.0 & 50.2 & 46.1 & 44.1 & 40.2 & 46.7 & 43.3 & 40.3 & 36.1 & 50.6 & 46.9 & 50.5 & 46.5 \\
Early Semantic Grounding-12B~\citep{he2026early} & 31.5 & 29.3 & 36.2 & 34.3 & 29.0 & 26.9 & 57.6 & 53.9 & 60.3 & 58.1 & 53.8 & 49.6 & 48.5 & 44.4 & 52.4 & 49.3 & 44.4 & 40.1 & 54.5 & 51.2 & 54.3 & 50.9 \\
TALENT~\citep{jin2026talent} & 59.1 & 57.0 & 63.7 & 61.8 & 56.6 & 54.4 & 77.8 & 75.9 & \cellcolor{TableRankYellow!15}79.4 & 78.3 & 74.8 & 72.8 & 70.1 & 66.9 & 74.9 & 72.3 & 63.4 & 58.8 & 69.7 & 65.9 & 69.2 & 66.8 \\
DeRIS-B~\citep{dai2025deris} & 73.9 & 68.0 & 72.5 & 71.6 & 65.2 & 63.1 & \cellcolor{TableRankOrange!30}81.1 & 78.4 & \cellcolor{TableRankOrange!30}81.9 & 79.8 & \cellcolor{TableRankOrange!30}79.5 & 77.6 & \cellcolor{TableRankOrange!30}74.3 & 70.1 & \cellcolor{TableRankOrange!30}77.9 & 75.4 & \cellcolor{TableRankOrange!30}71.2 & 67.5 & \cellcolor{TableRankGold!22}75.3 & 72.1 & \cellcolor{TableRankOrange!30}76.4 & 75.9 \\
\rowcolor{gray!15}
\contourvla-3B & \cellcolor{TableRankRed!38}\textbf{83.5} & \cellcolor{TableRankRed!38}\textbf{72.3} & \cellcolor{TableRankRed!38}\textbf{80.3} & \cellcolor{TableRankRed!38}\textbf{78.1} & \cellcolor{TableRankRed!38}\textbf{74.4} & \cellcolor{TableRankOrange!30}71.8 & \cellcolor{TableRankRed!38}\textbf{82.4} & \cellcolor{TableRankGold!22}81.3 & \cellcolor{TableRankRed!38}\textbf{83.7} & \cellcolor{TableRankYellow!15}82.9 & \cellcolor{TableRankRed!38}\textbf{81.1} & \cellcolor{TableRankOrange!30}79.2 & \cellcolor{TableRankRed!38}\textbf{77.5} & \cellcolor{TableRankGold!22}76.4 & \cellcolor{TableRankRed!38}\textbf{80.7} & \cellcolor{TableRankGold!22}80.7 & \cellcolor{TableRankRed!38}\textbf{74.2} & 69.3 & \cellcolor{TableRankRed!38}\textbf{76.6} & \cellcolor{TableRankGold!22}76.9 & \cellcolor{TableRankRed!38}\textbf{76.7} & \cellcolor{TableRankYellow!15}77.3 \\
\bottomrule
\end{tabular}}
\vspace{-1em}
\end{table}

\begin{figure}[htb]
    \centering
    \setlength{\abovecaptionskip}{-0.5em}
    \includegraphics[width=\linewidth]{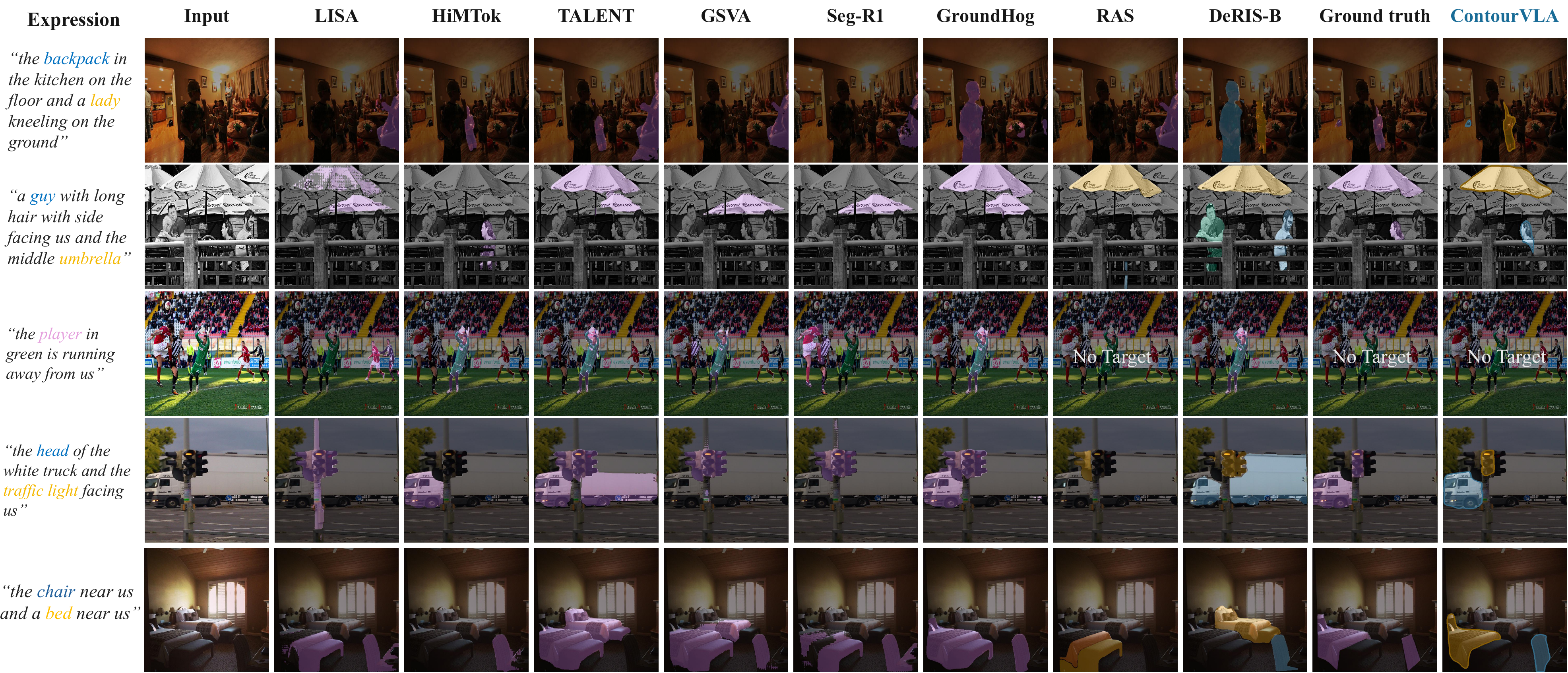}
    \caption{
    \textbf{Qualitative comparison on challenging GRES examples.}
    \contourvla\ recovers multiple and part-level referents, preserves fine
    boundaries, and correctly rejects the no-target case.
    }
    \label{fig:qualitative}
    \vspace{-1em}
\end{figure}

\textbf{Qualitative results.}
Figure~\ref{fig:qualitative} complements the quantitative evaluation with
multi-referent, part-level, and no-target examples. In the person-and-umbrella
example, \contourvla\ segments both requested referents; in the truck example,
it delineates the specified front portion rather than extending the mask over
the entire vehicle. It also correctly returns an empty prediction for the
absent-target expression. Compared with the displayed baselines, these examples
illustrate fewer omitted referents and less over-segmentation while preserving
fine boundaries. Together, they highlight two complementary requirements of
GRES: identifying the complete language-specified referent set and accurately
delineating the corresponding regions.

% \vspace{-0.5em}
\subsection{Ablation Studies and Analysis}
\label{sec:ablation}
\vspace{-0.5em}

\textbf{Closed-loop perception and EASS.}
Table~\ref{tab:eass_ablation} evaluates spatial sampling, semantic routing,
and re-observation on gRefCOCO val. All variants are trained separately under
the same SFT protocol, with identical grounding predictions, initial contours,
GAD configuration, and evolution budget. The baseline (A) uses vertex
sampling with a predefined VLIM layer schedule. Adding SVS (B) or MSR (C)
improves both metrics, while their combination (D) raises gIoU from 71.1 to
77.6 and cIoU from 60.8 to 67.4. Freezing the initial EASS readouts across later iterations (E),
while contour states and actions continue to evolve, reduces performance to
72.0 gIoU and 61.3 cIoU, supporting the benefit of state-conditioned
re-observation.

\textbf{Effectiveness of DECT-GRPO.}
Table~\ref{tab:dect_ablation} evaluates RL post-training and transport-derived
instance-level advantage modulation on gRefCOCO val. All RL variants start from
the same SFT checkpoint and share training data, rollout settings,
transport-based reward, and optimization budget. Shared-advantage GRPO sets
$\lambda_c=0$, while hard-assignment credit replaces soft transport only in
credit computation. Shared-advantage GRPO improves over SFT, while full
DECT-GRPO gains a further 3.0 gIoU and 5.1 instance F1@0.5 points, supporting
instance-level modulation beyond shared rollout feedback. Its gains over
hard-assignment credit (1.3 gIoU, 2.0 F1@0.5) further support soft correspondences.
Removing dustbins lowers all four metrics, particularly no-target accuracy and
instance F1@0.5, supporting dustbin-aware credit assignment for unmatched predictions and targets.

\vspace{-1em}
\begin{table}[H]
    \centering
    \begin{minipage}[t]{0.49\linewidth}
        \centering
        \caption{Ablation of EASS and re-observation on gRefCOCO val (\%).
        Refresh denotes EASS readout recomputation after contour updates.}
        \label{tab:eass_ablation}
        \renewcommand{\arraystretch}{1.08}
        \resizebox{\linewidth}{!}{%
        \begin{tabular}{lccccc}
            \toprule
            Variant & SVS & MSR & Refresh
            & gIoU $\uparrow$ & cIoU $\uparrow$ \\
            \midrule
            A: Baseline
            & $\times$ & $\times$ & $\checkmark$
            & 71.1 & 60.8 \\
            B: + Spatial sampling
            & $\checkmark$ & $\times$ & $\checkmark$
            & 74.2 & 63.5 \\
            C: + Semantic routing
            & $\times$ & $\checkmark$ & $\checkmark$
            & 74.7 & 63.1 \\
            \textbf{D: Full EASS}
            & $\checkmark$ & $\checkmark$ & $\checkmark$
            & \textbf{77.6} & \textbf{67.4} \\
            \midrule
            E: Frozen observations
            & $\checkmark$ & $\checkmark$ & $\times$
            & 72.0 & 61.3 \\
            \bottomrule
        \end{tabular}}
    \end{minipage}%
    \hfill
    \begin{minipage}[t]{0.49\linewidth}
        \centering
        \caption{Ablation of DECT-GRPO on gRefCOCO val (\%).
        N-acc is no-target accuracy; instance F1@0.5 uses one-to-one mask matching.}
        \label{tab:dect_ablation}
        \renewcommand{\arraystretch}{1.08}
        \resizebox{\linewidth}{!}{%
        \begin{tabular}{lcccc}
            \toprule
            Variant
            & gIoU $\uparrow$ & cIoU $\uparrow$
            & N-acc $\uparrow$ & F1@0.5 $\uparrow$ \\
            \midrule
            SFT only
            & 77.6 & 67.4
            & 74.6 & 73.8 \\
            \midrule
            GRPO (shared advantage)
            & 80.5 & 70.1
            & 78.0 & 77.3 \\
            Hard-assignment credit
            & 82.2 & 71.6
            & 80.9 & 80.4 \\
            Dustbin-free credit
            & 82.4 & 71.8
            & 78.7 & 78.6 \\
            \textbf{Full DECT-GRPO}
            & \textbf{83.5} & \textbf{72.3}
            & \textbf{83.2} & \textbf{82.4} \\
            \bottomrule
        \end{tabular}}
    \end{minipage}
    \vspace{-1em}
\end{table}

\begin{figure}[H]
    \centering
    \setlength{\abovecaptionskip}{-0.50em}
    \includegraphics[width=\linewidth]{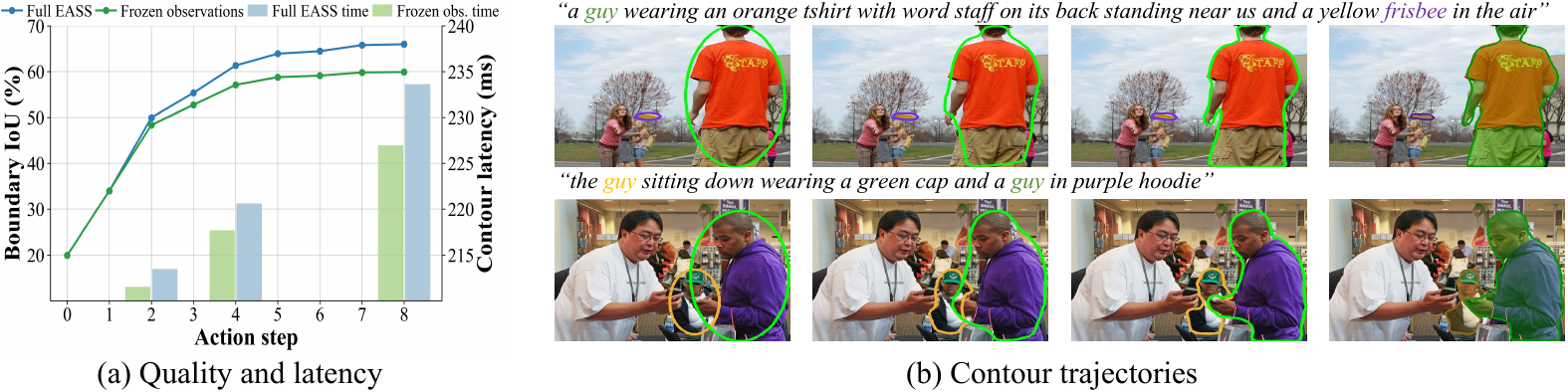}
    \caption{\textbf{Contour evolution and latency on a subset of gRefCOCO
    val.} (a) Boundary IoU (lines) and cumulative contour latency at steps
    2, 4, and 8 (bars) for Full EASS and frozen observations. (b) Full EASS
    evolution from initial contours to final masks, with
    consistent instance colors.}
    \label{fig:contour_evolution}
\end{figure}

\textbf{Contour evolution and computational cost.}
Figure~\ref{fig:contour_evolution}(a) compares Full EASS (D) and frozen
observations (E) under identical grounding predictions and initial contours,
reporting cumulative contour latency at action steps 2, 4, and 8. At the
matched eight-step action budget, Full EASS improves Boundary IoU by 6.07
points over frozen observations, from 59.91\% to 65.98\%.
Cumulative contour latency increases from 226.96 to 233.63\,ms, corresponding
to an additional 6.67\,ms (2.94\%). This comparison supports the benefit of
state-conditioned re-observation at matched action budgets with a modest
increase in contour-branch latency. Figure~\ref{fig:contour_evolution}(b)
visualizes the corresponding Full EASS evolution from initial contours to
final masks. Detailed evaluation and timing protocols are provided in
Appendix~\ref{app:training_eval}.

\FloatBarrier

\vspace{-1em}
\section{Conclusion}
\label{sec:conclusion}
% \vspace{-0.5em}

We presented ContourVLA, which recasts GRES as a closed-loop perception--action process in which editable contours guide multimodal perception and evolve through geometric actions. EASS couples contour-guided boundary sampling with multilevel semantic routing, while DECT-GRPO jointly optimizes grounding and contour actions through transport-based rewards and instance-level credit assignment. Among the compared methods, ContourVLA achieves the highest gIoU on all three gRefCOCO splits (by 2.7--8.7 points) and mIoU on all eight RES splits. Ablations further support the benefits of re-observation and instance-level credit, highlighting the value of coupling adaptive perception with explicit geometric refinement.

\begingroup
\raggedright
\bibliography{references}
\bibliographystyle{iclr2027_conference}
\endgroup

\clearpage
\appendix
\section{Additional Implementation and Experimental Details}

\subsection{Training and Evaluation Protocol}
\label{app:training_eval}

The complete training pipeline comprises grounding pretraining followed by the two-stage contour-policy training described in the main text: supervised fine-tuning (SFT) and DECT-GRPO post-training. Across these three chronological stages, \contourvla\ progressively acquires referent grounding, geometric contour control, and joint grounding-action optimization. The visual and language backbones are initialized from MoonViT and Qwen2.5-3B-Instruct, respectively.

\begin{figure}[H]
    \centering
    \includegraphics[width=0.90\linewidth]{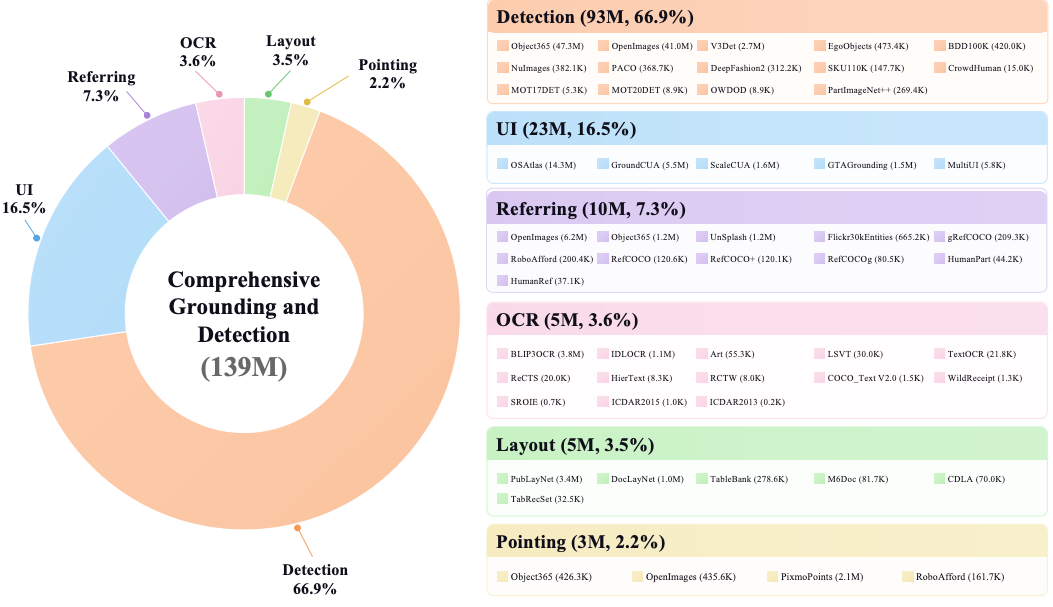}
    \caption{
    \textbf{Composition of the 139M LocateAnything-Data training instances used in Stage I.}
    The mixture contains detection (66.9\%), UI (16.5\%), referring (7.3\%),
    OCR (3.6\%), layout (3.5\%), and pointing (2.2\%) data.
    }
    \label{fig:data-distribution}
\end{figure}

\paragraph{Stage I: Grounding and Detection.}
We first train VLIM on LocateAnything-Data~\citep{wang2026locateanything}, a large-scale mixture of object detection and vision-language grounding tasks. Stage I contains 139M training instances; Figure~\ref{fig:data-distribution} summarizes their composition. Detection provides most of the supervision, while UI, referring, OCR, layout, and pointing data contribute complementary signals for fine-grained, language-conditioned localization. This stage adapts the initialized multimodal backbone to structured referent localization. VLIM learns to produce language-conditioned grounding outputs, including bounding boxes and no-target responses, while retaining the general visual-language representations inherited from pretraining. Because LocateAnything-Data primarily provides localization annotations, Stage I trains grounding and detection rather than contour actions.

\paragraph{Stage II: Supervised Contour Fine-Tuning.}

Starting from the Stage-I checkpoint, we perform SFT jointly on the training
sets of gRefCOCO~\citep{liu2023gres}, RefCOCO and
RefCOCO+~\citep{yu2016modeling}, and RefCOCOg~\citep{mao2016generation} to
initialize the contour-based perception-action policy. Ground-truth boxes
supervise VLIM grounding, and segmentation masks are converted into aligned
contour targets for GAD. Perturbed copies of the ground-truth boxes initialize
the contour-action branch, after which reference contour trajectories
supervise the evolution from these initial states toward the target
boundaries.

This stage jointly establishes referent grounding and geometric action prediction. Auxiliary geometric and routing objectives stabilize contour evolution and EASS optimization, as detailed in Appendix~\ref{sec:sft-losses}. The resulting checkpoint initializes DECT-GRPO post-training.

\paragraph{Stage III: DECT-GRPO Post-Training.}

The final stage applies DECT-GRPO to the same four-dataset training mixture,
initialized from the Stage-II checkpoint. For each image-expression pair,
VLIM samples discrete grounding outputs that define the initial contours, and
GAD then samples continuous geometric action trajectories. The final contours
are rasterized into segmentation masks and evaluated by dustbin-augmented
entropic transport.

The resulting transport plan provides both a transport-based rollout reward and instance-level credit maps. The reward is used to compute the shared group-relative advantage, whereas the credit maps assign instance-specific corrections to the corresponding grounding tokens and contour actions. DECT-GRPO therefore jointly optimizes the autoregressive grounding policy and the continuous contour-action policy from final segmentation feedback, without retaining the supervised contour losses used in Stage II.

\paragraph{Data splits and checkpoint evaluation.}
Stage I performs grounding and detection pretraining on
the 139M-instance LocateAnything-Data corpus. To prevent benchmark
information leakage, entries originating from gRefCOCO, RefCOCO, RefCOCO+,
or RefCOCOg are restricted to their official training splits; no validation
or test annotations or expressions from these benchmarks are used during
training. For Stages II and III, we likewise pool only the designated
training splits of the four benchmarks, retaining the same data mixture for
SFT and DECT-GRPO. After Stage III, a single checkpoint is evaluated on all
validation and test splits without further adaptation to individual
benchmarks.

\begin{table}[t]
\centering
\caption{\textbf{Complete training pipeline of \contourvla.} The
four-dataset mixture comprises the training sets of gRefCOCO, RefCOCO,
RefCOCO+, and RefCOCOg.}
\label{tab:training-protocol}
\small
\setlength{\tabcolsep}{5pt}
\renewcommand{\arraystretch}{1.16}
\begin{tabularx}{\linewidth}{@{}>{\raggedright\arraybackslash}p{0.18\linewidth}>{\raggedright\arraybackslash}p{0.23\linewidth}>{\raggedright\arraybackslash}X@{}}
\toprule
Stage & Training data & Supervision and objective \\
\midrule
\textbf{I. Grounding}
& LocateAnything-Data
& Boxes and grounding responses; comprehensive grounding and detection
pretraining. \\
\addlinespace[2pt]
\textbf{II. Contour SFT}
& Four-dataset mixture
& Boxes, masks, and aligned contour trajectories; joint grounding and
geometric-action learning. \\
\addlinespace[2pt]
\textbf{III. DECT-GRPO}
& Four-dataset mixture
& Final segmentation feedback; joint optimization of the grounding and
contour-action policies. \\
\bottomrule
\end{tabularx}
\end{table}

\paragraph{Mask-level metrics.}
Each image-expression pair is treated as one evaluation sample. Predicted
instance masks are merged by union, and the target mask is the union of all
instances referred to by the expression. For RES, mIoU averages foreground
IoU over the evaluated samples. For GRES, gIoU extends this average to
no-target samples~\citep{liu2023gres}: an empty prediction receives a score
of one when the target is empty, whereas a nonempty prediction receives zero.
An empty prediction for a nonempty target also receives zero. The cIoU metric
divides the summed intersection area by the summed union area over the
evaluated split. Correct empty predictions contribute zero to both sums,
while foreground predictions on no-target samples contribute to the union.

\paragraph{No-target accuracy.}
N-acc is the percentage of ground-truth no-target samples for which the
predicted mask contains no foreground pixels. Its denominator includes only
no-target samples, rather than the full evaluation set.

\paragraph{Instance-level evaluation.}
Instance F1@0.5 is computed from individual instance masks before union. For
each expression, we construct a one-to-one matching that maximizes the number
of prediction-target pairs with mask IoU of at least 0.5. Matched pairs are
true positives (TP); unmatched predictions and targets are false positives
(FP) and false negatives (FN), respectively. Counts are accumulated over all
evaluated expressions, and
\begin{equation}
\mathrm{F1@0.5}
=
\frac{2\sum \mathrm{TP}}
{2\sum \mathrm{TP}+\sum \mathrm{FP}+\sum \mathrm{FN}}.
\end{equation}
For no-target expressions, predicted instances contribute false positives,
while a correct empty prediction contributes no counts. For expressions with
targets but no predictions, all targets contribute false negatives. The final
score is reported as a percentage.

\paragraph{Boundary evaluation.}
For Figure~\ref{fig:contour_evolution}, Boundary IoU (bIoU)~\citep{cheng2021boundary}
is evaluated at the original image resolution on the same fixed subset of
gRefCOCO validation expressions containing targets. At each action step, the
predicted contours are rasterized and merged by union; the ground-truth
instance masks are merged in the same way. Boundary regions are extracted by
subtracting an eroded mask from the original mask, using a $3\times3$ kernel
with
\begin{equation}
r=\max\!\left(1,\operatorname{round}
\left(0.02\sqrt{H^2+W^2}\right)\right)
\end{equation}
erosion iterations, where $H$ and $W$ are the original image dimensions.
Zero padding accounts for boundaries touching the image border. Boundary IoU
is the intersection-over-union of the predicted and ground-truth boundary
regions, averaged over expressions and reported as a percentage. Samples with
no predicted mask receive zero and remain in the average at every step. This
evaluation operates on expression-level union masks and does not require
instance matching.

\paragraph{Latency protocol.}
For Figure~\ref{fig:contour_evolution}(a), Full EASS and frozen observations
use identical grounding predictions and initial contours, with two evolution
iterations and four action steps per iteration. Cumulative contour latency is
measured at action steps 2, 4, and 8 on the shared subset with predicted boxes.
The step-2 measurement includes the cost of predicting the complete first
four-step action chunk and executing its first two residual steps. Timing
includes contour-branch preparation, EASS readouts, GAD execution, and contour
updates, while excluding image encoding, grounding generation, and mask
rasterization. Intermediate contours are recorded from each model's complete
inference run.

\subsection{Additional Qualitative Comparisons}
\label{sec:additional-qualitative}

\begin{figure}[!t]
    \centering
    \includegraphics[pagebox=cropbox,width=\linewidth]{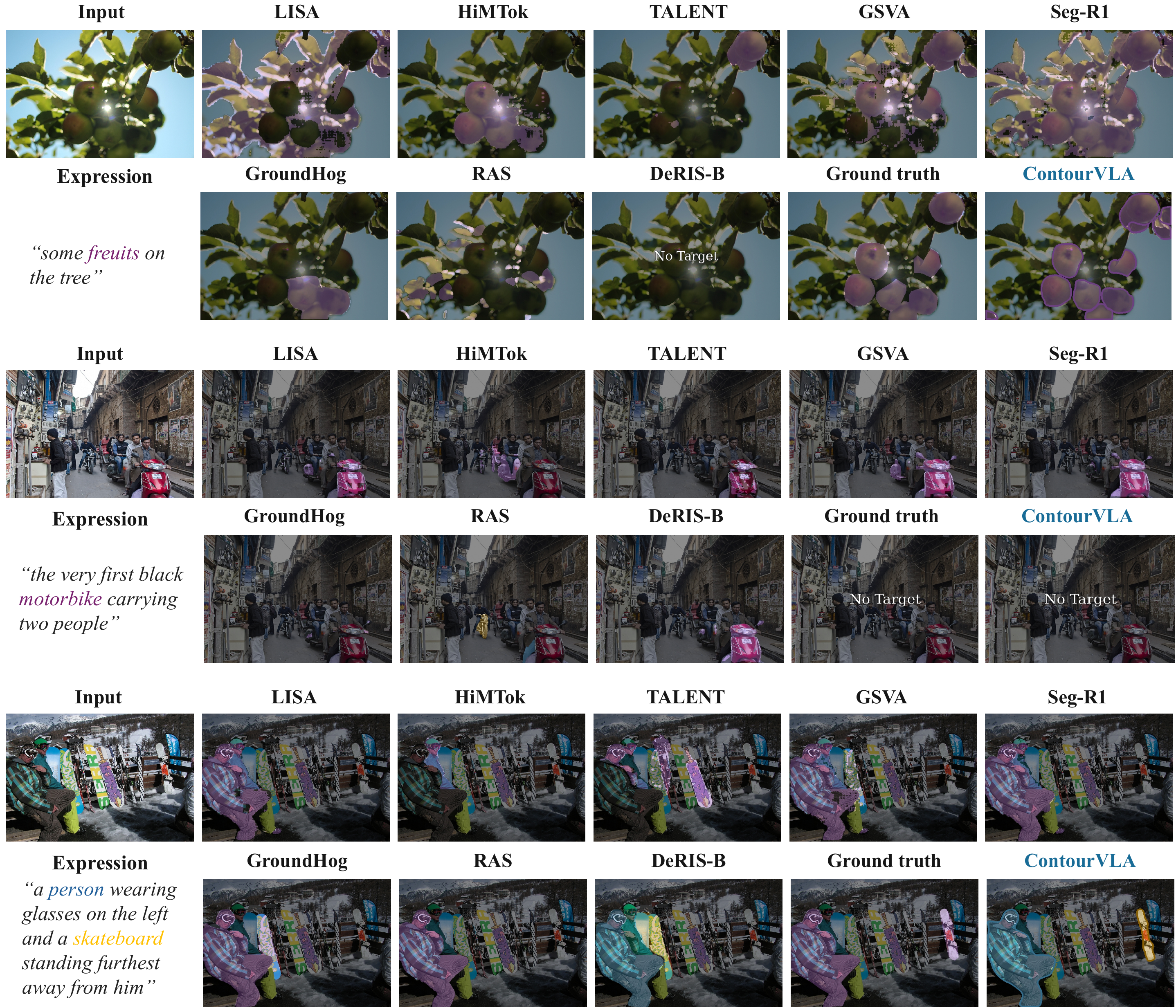}
    \caption{\textbf{Qualitative comparisons on plural, no-target, and
    relational referring expressions.}}
    \label{fig:supplementary-qualitative-1}
\end{figure}

\begin{figure}[!t]
    \centering
    \includegraphics[pagebox=cropbox,width=\linewidth]{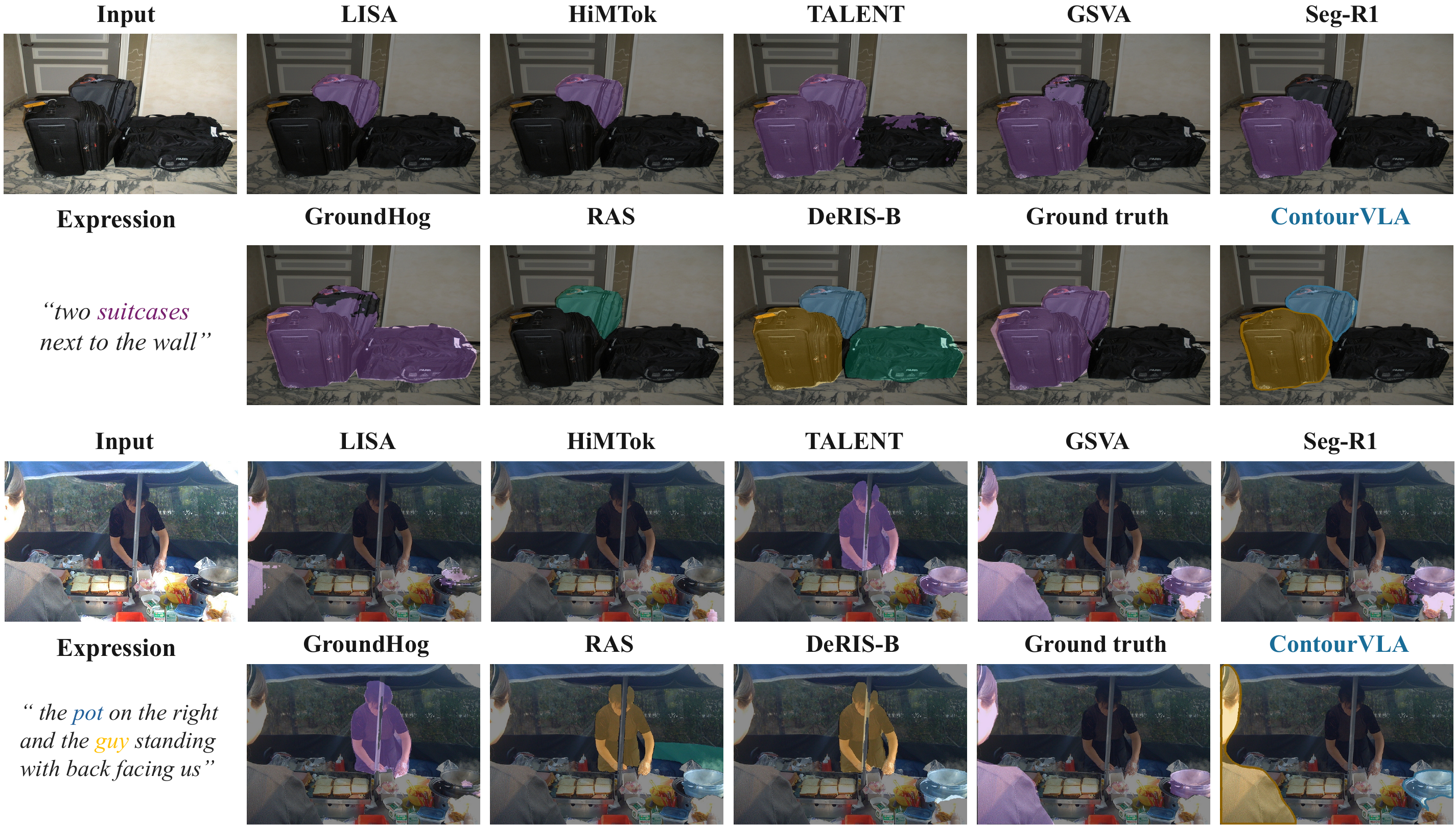}
    \caption{\textbf{Qualitative comparisons on numerical, spatial, and
    heterogeneous queries.}}
    \label{fig:supplementary-qualitative-2}
\end{figure}

Figures~\ref{fig:supplementary-qualitative-1}
and~\ref{fig:supplementary-qualitative-2} provide additional side-by-side
comparisons on challenging referring expressions, complementing the
quantitative evaluation in the main text.

Figure~\ref{fig:supplementary-qualitative-1} examines plural fruit referents,
an absent target, and a multi-referent expression combining visual attributes
and spatial relations. Different mask colors distinguish referred instances,
while ``No Target'' denotes an empty prediction. These examples jointly test
referent-set completeness, false-positive suppression, and boundary quality.

Figure~\ref{fig:supplementary-qualitative-2} focuses on numerical and spatial
constraints for multiple suitcases and on a heterogeneous object--person query
involving a pot and a person. Together with Figure~\ref{fig:supplementary-qualitative-1},
it covers plural and multi-instance descriptions, heterogeneous referents, and
the no-target case.

\subsection{Out-of-Distribution Video Segmentation}
\label{app:ood-video}

\begin{figure}[!t]
    \centering
    \begin{minipage}[t]{0.49\linewidth}
        \centering
        \includegraphics[width=\linewidth]{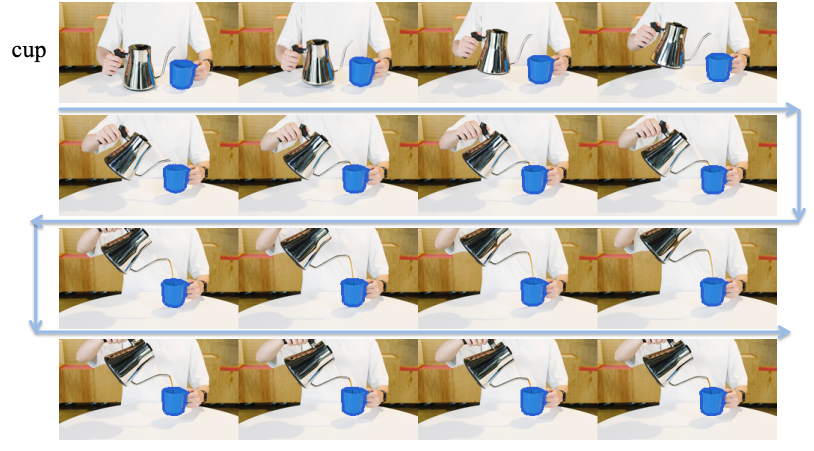}\\[-0.25em]
        \small (a)
    \end{minipage}%
    \hfill
    \begin{minipage}[t]{0.49\linewidth}
        \centering
        \includegraphics[width=\linewidth]{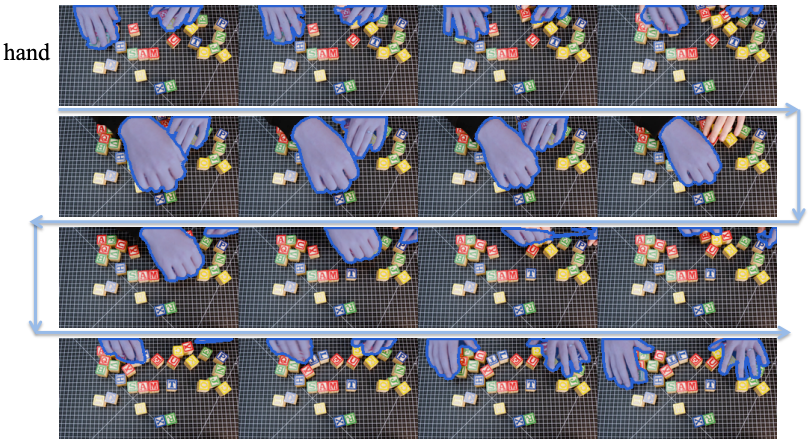}\\[-0.25em]
        \small (b)
    \end{minipage}
    \vspace{0.35em}

    \begin{minipage}[t]{0.49\linewidth}
        \centering
        \includegraphics[width=\linewidth]{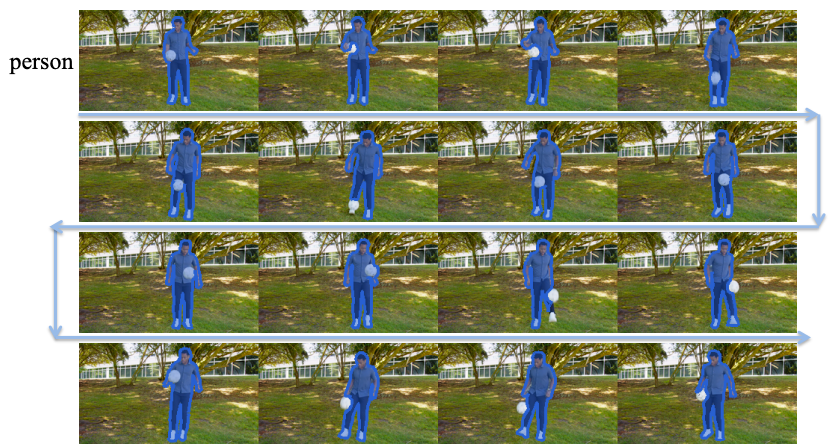}\\[-0.25em]
        \small (c)
    \end{minipage}%
    \hfill
    \begin{minipage}[t]{0.49\linewidth}
        \centering
        \includegraphics[width=\linewidth]{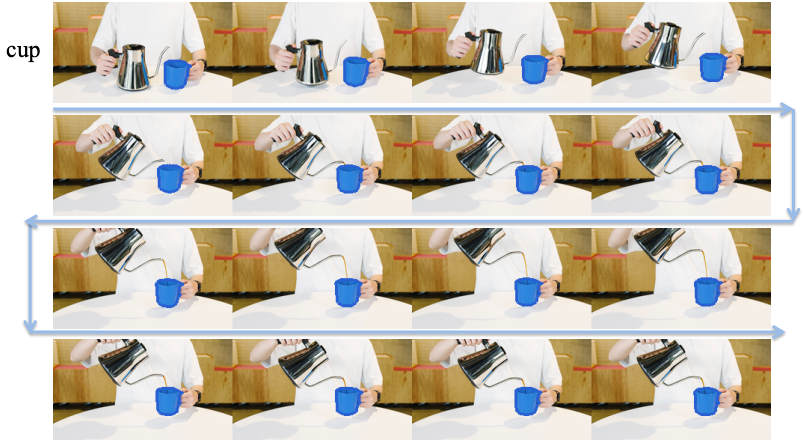}\\[-0.25em]
        \small (d)
    \end{minipage}
    \caption{\textbf{Qualitative segmentation on out-of-distribution videos.}
    Panels (a)--(d) show sampled frames in temporal order, indicated by the
    arrows. The referring query is displayed at the left of each panel. }
    \label{fig:ood-video}
\end{figure}

To examine qualitative robustness beyond the image benchmarks, we apply the
trained \contourvla\ checkpoint to sampled frames from four out-of-distribution
video examples. Figure~\ref{fig:ood-video} shows predictions in temporal order
for the query displayed at the left of each panel. Across object manipulation,
hand--object interaction, and human motion, the predicted masks generally
remain aligned with the referred regions despite changes in pose, position,
appearance, and partial occlusion. This experiment provides a qualitative OOD
assessment and does not report video-level temporal-consistency metrics.

\subsection{State Descriptors and Decoder Initialization}
\label{sec:implementation-details}

\paragraph{Seven-dimensional routing state.}
The implementation instantiates the factors $(g_i,c_i^k,r^k)$ in \eqref{eq:semantic-query} with a seven-dimensional state descriptor: log box aspect ratio, log box area, normalized point allocation, contour irregularity, normalized evolution index, and the mean and maximum displacement from the preceding evolution iteration. Contour irregularity is the coefficient of variation of the vertex radii about the box center in box-relative coordinates. Displacements are also normalized by box width and height. All statistics exclude padded vertices, and the two displacement entries are set to zero at the initial iteration. For fixed boxes and point allocations, the first three descriptors remain constant across iterations and encode instance characteristics rather than evolution progress.

\paragraph{Initialization and routing regularization.}
Within $f_q$, the final projection of the state-dependent component is zero-initialized. Consequently, the initial query is determined by the evolution-depth slot embedding, while the state-dependent offset is introduced gradually during training. The output projection of the residual semantic-attention branch is also zero-initialized, so the initial GAD block preserves its spatial and instance-context pathway. During SFT, aggregate-usage, maximum-probability, and expected-depth regularizers constrain the router as specified in \eqref{eq:sft-router}. During DECT-GRPO, EASS is optimized through the action likelihood and the straight-through routing path, without an additional router loss.

\paragraph{Routing-score scale.}
In \eqref{eq:state-routing}, normalized queries and keys yield dot products in $[-1,1]$. When the prior weight is zero, the largest soft routing probability is bounded by
\begin{equation}
\max_{\ell}p_{i,\ell}^{k,d}\leq
\frac{\exp(2)}{\exp(2)+|\mathcal L|-1}.
\label{eq:routing-confidence-bound}
\end{equation}
The forward argmax still selects a single layer, but the bound limits the attainable value of a confidence objective defined on the soft probabilities. Equation~\ref{eq:state-routing} uses neither a routing temperature nor an additional logit scale.

\subsection{Coordinate Conventions and Action-Chunk Execution}
\label{sec:coordinate-action-details}

\paragraph{Box-normalized normal-direction sampling.}
All image coordinates are normalized to $[0,1]^2$. For box
$b_i=(x_i^1,y_i^1,x_i^2,y_i^2)$, define its origin
$q_i=(x_i^1,y_i^1)$ and anisotropic scale matrix
\begin{equation}
D_i=\operatorname{diag}(w_i,h_i),\qquad
\overline c_{i,p}^k=D_i^{-1}(c_{i,p}^k-q_i).
\label{eq:box-coordinate-transform}
\end{equation}
The cyclic central-difference tangent and outward unit normal are computed
in this box-normalized coordinate system:
\begin{equation}
\overline\tau_{i,p}^k=
\overline c_{i,p+1}^k-\overline c_{i,p-1}^k,\qquad
n_{i,p}^k=
\frac{R_{\mathrm{out}}\overline\tau_{i,p}^k}
{\max(\|\overline\tau_{i,p}^k\|_2,10^{-6})},
\label{eq:box-normal}
\end{equation}
where $R_{\mathrm{out}}$ rotates the tangent by $-\pi/2$ for a
counterclockwise contour and by $+\pi/2$ for a clockwise contour. Indices
are cyclic, and the contour orientation is determined from its signed area.
The sampling coordinates used in \eqref{eq:normal-observation} are therefore
\begin{equation}
s_{i,p,j}^{k,\pm}
=c_{i,p}^k\pm o_jD_i n_{i,p}^k
=c_{i,p}^k\pm o_j
\bigl(n_{i,p}^k\odot(w_i,h_i)\bigr).
\label{eq:box-normal-sampling}
\end{equation}
This is strict normal sampling in the normalized-box domain and
box-scale-normalized directional sampling in the image domain. Under
anisotropic scaling, $D_i n_{i,p}^k$ need not be parallel to the Euclidean
image-space normal; the latter would instead be proportional to
$D_i^{-\top}n_{i,p}^k$ and is not used here. Before bilinear interpolation,
$s=(s_x,s_y)$ is mapped to feature-grid coordinates
$((W_F-1)s_x,(H_F-1)s_y)$ and clipped to the valid grid boundary.

\paragraph{Chunk prediction and state transition.}
At closed-loop iteration $k$, EASS is evaluated once on
$C_i^{k,0}=C_i^k$. Conditioned on this observation, GAD predicts the entire
chunk
$A_i^k=\{a_{i,p}^{k,t}\}_{p=1,t=1}^{P_i,T}$ in parallel. Each bounded
action is a residual in box-normalized coordinates:
\begin{equation}
a_{i,p}^{k,t}=\delta_{\max}\tanh(u_{i,p}^{k,t}),\qquad
\overline c_{i,p}^{k,t}
=\overline c_{i,p}^{k,t-1}+a_{i,p}^{k,t},
\label{eq:normalized-action-update}
\end{equation}
or, equivalently,
\begin{equation}
c_{i,p}^{k,t}
=c_{i,p}^{k,t-1}+D_i a_{i,p}^{k,t}.
\label{eq:image-action-update}
\end{equation}
The $T$ residual fields are executed sequentially without refreshing visual
or semantic observations inside the chunk. Coordinates are clipped to the
image domain, while vertex ordering and valid-point counts remain fixed.
After the final within-chunk step,
$C_i^{k+1}=C_i^{k,T}$; normals, spatial samples, routing queries, and GAD
observations are then recomputed before predicting the next chunk. Thus,
$T$ controls the open-loop action horizon, whereas $K$ controls the number
of perception-action cycles. Deterministic SFT and inference use
$u=\mu$; DECT-GRPO samples $u$ according to
\eqref{eq:action-logprob}.

\paragraph{Centralized implementation configuration.}
Table~\ref{tab:implementation-config} reports the default geometry and
optimization settings. With $K=2$ and $T=4$, the policy executes eight
action steps and re-observes the contour after every four steps. Since
$\delta_{\max}=0.05$, one step can move a vertex by at most $5\%$ of the
box width and height along the two normalized coordinate axes.

\begin{table}[!t]
\centering
\small
\caption{Default spatial sampling, action execution, and optimization
configuration. Stage-II and Stage-III batch sizes are global.}
\label{tab:implementation-config}
\begin{tabularx}{\linewidth}{@{}l>{\raggedright\arraybackslash}X>{\raggedright\arraybackslash}X@{}}
\toprule
Component & Setting & Interpretation \\
\midrule
Input / feature grid
& $896{\times}896$ / $56{\times}56$
& Fixed-resolution input and stride-16 VLIM grid \\
Contour vertices
& $P_i\in[64,128]$
& Perimeter-adaptive valid-point count \\
Spatial offsets
& $J=4$, $\{o_j\}=\{1/64,1/32,1/16,1/8\}$
& Fractions of box-normalized distance on each side \\
GAD depth
& $D=6$ blocks
& One routed semantic feature per block \\
Action horizon
& $K=2$ chunks, $T=4$ steps/chunk
& Eight action steps, with re-observation between the two chunks. \\
Action bound
& $\delta_{\max}=0.05$
& Per-axis bound in normalized-box coordinates \\
\midrule
VLIM adaptation
& LoRA rank 64, $\alpha=128$, dropout $0.05$
& Applied to attention Q/K/V/O and MLP gate/up/down projections \\
Stage-II SFT
& 30k steps; batch 128; LR $2{\times}10^{-5}$ (LoRA) and
$1{\times}10^{-4}$ (EASS/GAD)
& Four-dataset joint supervised training \\
Stage-III DECT-GRPO
& 8k updates; 32 input groups/update; $G=8$; LR
$5{\times}10^{-6}$ (LoRA) and $2{\times}10^{-5}$ (EASS/GAD)
& One update per sampled rollout group \\
Grounding sampling
& temperature $[0.65,0.75]$, top-$p=0.95$
& Grouped rollout diversity with fixed nucleus threshold \\
Optimizer
& AdamW, $(\beta_1,\beta_2)=(0.9,0.95)$, weight decay $0.01$
& Cosine decay, $3\%$ warmup, gradient clipping at $1.0$ \\
Numerics
& bfloat16 with float32 rewards and transport
& Mixed-precision policy optimization \\
\bottomrule
\end{tabularx}
\end{table}

\subsection{Supervised Targets and Loss Details}
\label{sec:sft-losses}

\paragraph{Grounding targets and initial contours.}
SFT uses teacher-forced grounding answers constructed from quantized and subsequently dequantized ground-truth boxes. A separate copy of each box is perturbed to initialize the contour-action branch; the grounding targets themselves are not jittered. Each perturbed box initializes an elliptical contour with 64-128 valid vertices, and the corresponding ground-truth boundary is resampled by arc length to the same number of vertices. The perturbation magnitude increases linearly from $0.05$ to $0.10$ during the first 3,000 optimization steps.

\paragraph{Reference contour trajectories.}
Let $\Gamma_i$ be the resampled ground-truth contour. We align it to $C_i^0$ by minimizing squared vertex distances over cyclic shifts and both traversal directions, denoted by $\mathcal P_{\mathrm{cyc}}$:
\begin{equation}
\pi_i^\star=\arg\min_{\pi\in\mathcal P_{\mathrm{cyc}}}
\sum_{p=1}^{P_i}\|\Gamma_{i,\pi(p)}-c_{i,p}^{0}\|_2^2,
\qquad c_{i,p}^\star=\Gamma_{i,\pi_i^\star(p)}.
\label{eq:sft-alignment}
\end{equation}
This operation fixes the within-instance vertex correspondence and is distinct from the inter-instance transport used by DECT-GRPO. At evolution iteration $k$, the reference position for action step $t$ interpolates from the iteration's actual starting contour to the aligned target:
\begin{equation}
\widetilde C_i^{k,t}=\operatorname{sg}(C_i^k)
+\frac{t}{T}\bigl(C_i^\star-\operatorname{sg}(C_i^k)\bigr).
\label{eq:sft-reference}
\end{equation}
The references in \eqref{eq:sft-reference} provide soft positional
supervision through \eqref{eq:sft-trajectory}, rather than executable
expert-action trajectories. The predicted contour remains subject to the
bounded transitions in
\eqref{eq:normalized-action-update}-\eqref{eq:image-action-update}, so these
references are not guaranteed to be reachable within every chunk. Reaching
the aligned target at each chunk endpoint is therefore a supervision
objective, not a hard constraint.
The stop-gradient operator is applied only when constructing the reference. The predicted contour trajectory itself remains connected across iterations, so later trajectory losses can backpropagate through earlier contour updates.

\paragraph{Smooth-$\ell_1$ and trajectory weights.}
For a scalar residual, the smooth-$\ell_1$ penalty is
\begin{equation}
h_\beta(r)=
\begin{cases}
r^2/(2\beta), & |r|<\beta,\\
|r|-\beta/2, & \text{otherwise},
\end{cases}
\qquad \beta=0.05.
\label{eq:smooth-l1}
\end{equation}
With $k=0,\ldots,K-1$ and $t=1,\ldots,T$, the time weights in \eqref{eq:sft-trajectory} are
\begin{equation}
\omega_{k,t}=\frac{((kT+t)/(KT))^\gamma}
{\sum_{k'=0}^{K-1}\sum_{t'=1}^{T}((k'T+t')/(KT))^\gamma},
\qquad\gamma=1.
\label{eq:sft-time-weights}
\end{equation}
These weights sum to one over all evolution iterations and action steps. To emphasize vertices that must travel farther, we define box-normalized distances from the initial contour:
\begin{equation}
\begin{aligned}
d_{i,p}&=\|(c_{i,p}^\star-c_{i,p}^{0})\oslash b_i^{\mathrm{size}}\|_2,
\qquad \overline d_i=P_i^{-1}\sum_p d_{i,p},\\
v'_{i,p}&=1+\lambda_{\mathrm{travel}}\frac{d_{i,p}}{\max(\overline d_i,\epsilon_{\mathrm{num}})},
\qquad
v_{i,p}=\frac{v'_{i,p}}{P_i^{-1}\sum_{q=1}^{P_i}v'_{i,q}},
\end{aligned}
\label{eq:sft-travel-weights}
\end{equation}
where $\lambda_{\mathrm{travel}}=1$ and $\epsilon_{\mathrm{num}}>0$ prevents division by zero. All averages exclude padded vertices. Normalizing $v_{i,p}$ to unit mean preserves the average weight of each instance, and \eqref{eq:sft-trajectory} weights instances equally rather than in proportion to their vertex counts.

\paragraph{Trajectory supervision.}
Let $\mathcal I$ contain instances with valid target contours. Their box-normalized residual and trajectory loss are
\begin{equation}
\begin{aligned}
r_{i,k,t,p}&=(c_{i,p}^{k,t}-\widetilde c_{i,p}^{k,t})\oslash b_i^{\mathrm{size}},\\
\mathcal L_{\mathrm{traj}}&=
\frac1{|\mathcal I|}\sum_{i\in\mathcal I}\frac1{P_i}
\sum_{p=1}^{P_i}v_{i,p}\sum_{k,t}\omega_{k,t}
\sum_{d\in\{x,y\}}h_{0.05}(r_{i,k,t,p,d}).
\end{aligned}
\label{eq:sft-trajectory}
\end{equation}
Box dimensions are lower-bounded before normalization. The temporal weights emphasize later actions, while the unit-mean travel weights emphasize vertices farther from their aligned targets.

\paragraph{Region supervision.}
At the end of each evolution iteration, we differentiably rasterize the predicted contour within an expanded region of interest (ROI) and compare it with the ground-truth mask rasterized in the same ROI:
\begin{equation}
\mathcal L_{\mathrm{region}}=
\frac{\sum_{k=0}^{K-1}\eta_k\operatorname{Mean}_{i\in\mathcal I}
\left[1-\operatorname{SoftIoU}\!\left(\mathcal R_{\mathrm{soft}}(C_i^{k,T}),Y_i^{\mathrm{ROI}}\right)\right]}
{\sum_{k=0}^{K-1}\eta_k},
\qquad\eta_k=\frac{k+1}{K}.
\label{eq:sft-region}
\end{equation}
The soft-rasterization temperature is $0.0025$. Unlike the mask evaluation used to construct the RL reward, this operation is differentiable with respect to the predicted contour.

\paragraph{Final-contour regularization.}
Let $X_{i,p}=c_{i,p}^{K}\oslash b_i^{\mathrm{size}}$ and use cyclic adjacency within each instance's valid point count. Smoothness penalizes deviation from the midpoint of neighboring vertices:
\begin{equation}
\mathcal L_{\mathrm{smooth}}=
\operatorname{Mean}_{i,p}\left\|X_{i,p}-\frac{X_{i,p-1}+X_{i,p+1}}2\right\|_1.
\label{eq:sft-smoothness}
\end{equation}
With $l_{i,p}=\|X_{i,p+1}-X_{i,p}\|_2$ and $\overline l_i=P_i^{-1}\sum_p l_{i,p}$, the spacing penalty is
\begin{equation}
\mathcal L_{\mathrm{spacing}}=
\operatorname{Mean}_i\frac1{P_i}\sum_{p=1}^{P_i}(l_{i,p}-\overline l_i)^2.
\label{eq:sft-spacing}
\end{equation}
This term promotes uniform edge lengths within each contour without imposing a common absolute spacing across objects. Both regularizers are applied only to the final contour.

\paragraph{Auxiliary signed-distance supervision.}
For locations $q$ on a $16\times16$ grid in each object ROI, let $d_i^\star(q)=\operatorname{SDF}(q,C_i^\star)$ denote the target signed distance. We normalize it by $r_i=\max((w_i+h_i)/2,10^{-3})$ and truncate only the target:
\begin{equation}
\begin{aligned}
\widetilde d_i(q)&=\operatorname{clip}\!\left(d_i^\star(q)/r_i,-\tau_{\mathrm{sdf}},\tau_{\mathrm{sdf}}\right),
\qquad\tau_{\mathrm{sdf}}=0.25,\\
\mathcal L_{\mathrm{SDF}}&=\operatorname{Mean}_{i,q}
 h_\beta\!\left(\widehat d_i(q)/r_i-\widetilde d_i(q)\right).
\end{aligned}
\label{eq:sft-sdf}
\end{equation}
The predicted distance is not clipped, which preserves gradients outside the target truncation interval.

\paragraph{Grounding supervision.}
Let $\mathcal S$ collect supervised answer-token positions. The teacher-forced grounding objective is
\begin{equation}
\mathcal L_{\mathrm{ground}}=-\frac1{|\mathcal S|}\sum_{n\in\mathcal S}
\log\pi_\theta^y(y_n^\star\mid I,x,y_{<n}^\star).
\label{eq:sft-grounding}
\end{equation}
Supervised positions include coordinate tokens, structural tags, no-target responses, and the end-of-sequence token. Prompt positions are excluded, and copied expression tokens inside \texttt{<ref>} tags are masked by default. The target coordinates are obtained from the annotated boxes before geometric perturbation. If an example contains no valid target contour, the geometric losses provide no contour-learning signal, whereas grounding supervision, including no-target responses, remains active.

\paragraph{Routing regularization during SFT.}
Index decoder slots first by evolution iteration $k=0,\ldots,K-1$ and then by block depth $d=1,\ldots,D$, with $s=kD+d\in\{1,\ldots,KD\}$, and write $p_{i,s,\ell}$ for their layer probabilities. Define $\overline p_\ell=\operatorname{Mean}_{i,s}p_{i,s,\ell}$ and expected layer depth $m_{i,s}=\sum_\ell\ell p_{i,s,\ell}$. The regularizer is
\begin{equation}
\begin{aligned}
\mathcal L_{\mathrm{router}}={}&-\lambda_{\mathrm{div}}H(\overline p)
-\lambda_{\mathrm{conf}}\operatorname{Mean}_{i,s}\max_\ell p_{i,s,\ell}\\
&+\lambda_{\mathrm{mono}}\operatorname{Mean}_{i,\,1\leq s<KD}
\operatorname{ReLU}(m_{i,s}-m_{i,s+1}).
\end{aligned}
\label{eq:sft-router}
\end{equation}
The first term encourages diverse aggregate layer usage; the second rewards confident selection through the maximum probability rather than by minimizing per-slot entropy. The final term penalizes decreases in expected layer depth between adjacent evolution--depth slots. The confidence objective respects the finite probability range in \eqref{eq:routing-confidence-bound} and therefore does not require probabilities to approach one. These auxiliary terms are used only during SFT and are not added separately during DECT-GRPO.

\paragraph{Effective SFT coefficients.}
The SFT objective used in our experiments is
\begin{equation}
\begin{aligned}
\mathcal L_{\mathrm{SFT}}={}&
\mathcal L_{\mathrm{ground}}+2\mathcal L_{\mathrm{traj}}
+0.04\mathcal L_{\mathrm{smooth}}+0.10\mathcal L_{\mathrm{spacing}}\\
&+0.70\mathcal L_{\mathrm{region}}+0.178\mathcal L_{\mathrm{SDF}}
+\mathcal L_{\mathrm{router}}.
\end{aligned}
\label{eq:sft-total}
\end{equation}
The coefficients internal to $\mathcal L_{\mathrm{router}}$ are specified separately. These supervised objectives initialize the policy and are not retained during DECT-GRPO post-training.

\subsection{DECT-GRPO Replay and Training Protocol}
\label{sec:joint-rl-replay}

\paragraph{Configuration and group-relative advantages.}
The DECT-GRPO configuration uses $G=8$, $\tau=0.001$, $\sigma_a=0.05$,
$\lambda_c=0.5$, $\lambda_a=1.0$, clipping range $c=0.2$,
$\beta_{\mathrm{KL}}=0.02$, transport entropy $\varepsilon=0.05$, and null
score $b_\varnothing=0.25$. For the $G$ rewards associated with one
image--expression pair, define
\begin{equation}
\overline R=\frac1G\sum_{g=1}^{G}R_g,\qquad
\sigma_R=\sqrt{\frac1G\sum_{g=1}^{G}(R_g-\overline R)^2},\qquad
A_g=\mathbf 1\{\sigma_R\geq\tau\}
\frac{R_g-\overline R}{\max(\sigma_R,\tau)}.
\label{eq:group-advantage}
\end{equation}
The threshold $\tau$ suppresses numerically unstable updates when the rewards within a group are effectively identical.

\paragraph{Rollout likelihoods and replay.}
For a sampled grounding sequence, the grounding branch follows an
autoregressive categorical policy with token-level log-probability
\begin{equation}
\ell^y_{g,n}
=
\log\pi_\theta^y(y_{g,n}\mid I,x,y_{g,<n}).
\label{eq:grounding-logprob}
\end{equation}
For each valid action index $\xi=(i,k,t,p)$, GAD predicts a two-dimensional
mean $\mu_{g,\xi}$ and samples
$u_{g,\xi}\sim\mathcal N(\mu_{g,\xi},\sigma_a^2I_2)$. The bounded contour
action is $a_{g,\xi}=\delta_{\max}\tanh(u_{g,\xi})$, with log-probability
\begin{equation}
\ell^a_{g,\xi}
=
\log\mathcal N(u_{g,\xi};\mu_{g,\xi},\sigma_a^2I_2)
-
\sum_{h=1}^{2}
\log\!\left[
\delta_{\max}\left(1-\tanh^2u_{g,\xi,h}\right)
\right].
\label{eq:action-logprob}
\end{equation}
During rollout collection, we store the sampled grounding outputs, pre-$\tanh$
action latents, behavior-policy log-probabilities, valid-point masks,
grounding boxes, and historical contour states. During replay, the sampled
quantities remain fixed, while current grounding log-probabilities,
trainable observations, and action means are recomputed with gradients
enabled. Dropout is disabled during both collection and replay.

\paragraph{Direct geometric policy gradients.}
For a fixed latent $u$ and fixed variance, the transformation Jacobian is constant with respect to the current mean. Thus
\begin{equation}
\frac{\partial\ell^a}{\partial\mu}=\frac{u-\mu}{\sigma_a^2},
\qquad
\nabla_\phi\mathcal L^a_g\approx
-\frac1{|\mathcal D_g|}\sum_{\xi\in\mathcal D_g}
A^a_{g,\xi}\nabla_\phi\ell^a_{g,\xi}.
\label{eq:action-score-gradient}
\end{equation}
This approximation holds when the likelihood ratios are close to one and clipping is inactive. The sampled continuous actions and their historical states are detached, while the current mean network remains differentiable. The resulting score-function gradient trains GAD and its observation pathways without differentiating through the mask reward. DECT-GRPO retains none of the SFT trajectory, region, smoothness, spacing, SDF, or router losses, and it introduces no action-reference KL or action-entropy term. Inference and SFT use deterministic actions $\delta_{\max}\tanh(\mu)$.

\paragraph{Output-count weighting.}
Grounding-token counts and valid-action counts can induce different centered corrections for the same instance. With correctly aligned, disjoint grounding spans and valid-point masks, each branch preserves its rollout-level advantage sum under instance-level modulation:
\begin{equation}
\sum_{n=1}^{L_g}A^y_{g,n}=L_gA_g,
\qquad
\sum_{\xi\in\mathcal D_g}A^a_{g,\xi}=|\mathcal D_g|A_g.
\label{eq:advantage-conservation}
\end{equation}
This conservation property does not make the branch gradients equal, because their likelihood models and normalizations differ. Within an instance, all valid action steps and vertices receive the same instance-corrected advantage; DECT-GRPO does not estimate an additional temporal credit signal. For a single predicted instance, branch-wise centering yields zero instance correction, but the rollout still receives the shared advantage $A_g$.

\paragraph{Grounding sampling and reference convention.}
Grounding answers use temperature and nucleus sampling. For each rollout, the temperature is sampled from $[0.65,0.75]$, while the nucleus threshold remains fixed. Replay scores the re-tokenized answer under the model's untruncated softmax. Because collection and replay therefore use different token distributions, the resulting objective should be interpreted as the implemented branch-wise policy-gradient surrogate rather than as an exactly unbiased on-policy estimator of the transformed grounding sampler. The grounding reference branch is formed by disabling the LoRA adapters while sharing the current visual-to-language projector. Reference log-probabilities are detached, but the reference is not a separately frozen copy of the complete SFT model. The exponential reference penalty is applied only to grounding and is not treated as an exact KL estimate under the transformed sampler. Behavior-policy log-probabilities are retained for both branches, and each rollout group is used for a single update rather than for multiple PPO epochs.

\subsection{Derivation of Dustbin-Augmented Entropic Transport}
\label{sec:dect-transport-derivation}

For rollout $g$, DECT solves the entropy-regularized transport problem
\begin{equation}
P_g^\star=
\arg\max_{P\in\mathcal U_g}
\left[
\langle P,\overline Q_g\rangle
-\varepsilon\sum_{i,j}P_{ij}\log P_{ij}
\right],
\label{eq:appendix-dect-objective}
\end{equation}
where $\mathcal U_g=\{P\geq0\mid P\mathbf 1=\mathbf a_g,\;
P^\top\mathbf 1=\mathbf b_g\}$ specifies the prediction-side and target-side marginals, with $\mathbf 1^\top\mathbf a_g=\mathbf 1^\top\mathbf b_g$. The augmented score matrix $\overline Q_g$ contains both real prediction--target IoUs and dustbin scores, so matched and unmatched instances are handled within a single transport problem.

\paragraph{Operational marginals, dustbin capacity, and solver settings.}
For $M_g>0$ and $N>0$, let $T_g=M_g+N$. We use the normalized
SuperGlue-style marginals
\begin{equation}
\mathbf a_g=
\frac{1}{T_g}
\begin{bmatrix}\mathbf 1_{M_g}\\ N\end{bmatrix},
\qquad
\mathbf b_g=
\frac{1}{T_g}
\begin{bmatrix}\mathbf 1_N\\ M_g\end{bmatrix}.
\label{eq:dect-marginals}
\end{equation}
Thus, every real prediction supplies $1/T_g$ mass and every real target
demands $1/T_g$ mass. The null-prediction row has capacity $N/T_g$, enough
to explain all targets as missed, while the null-target column has capacity
$M_g/T_g$, enough to absorb all predictions as unmatched. The null--null
entry receives whichever part of these capacities remains after real
matches are formed. Consequently, dustbin capacity is determined by
\eqref{eq:dect-marginals}, rather than merely by appending a row and column.

Mask IoUs are kept on their native $[0,1]$ scale. In all experiments we fix
$\varepsilon=0.05$ and $b_\varnothing=0.25$; hence a weak real match competes
directly with a null match of score $0.25$, while entropy smooths competition
within a scale of $0.05$. We perform Sinkhorn scaling in the log domain,
initialized with zero log-scalings, until the maximum row/column marginal
residual is below $10^{-6}$ or 1000 iterations are reached. Cases with
$M_g=0$ or $N=0$ bypass transport and use the explicit set rewards in
Section~\ref{sec:rl-protocol}.

\paragraph{Dustbin-free credit ablation.}
For the variant in Table~\ref{tab:dect_ablation}, the rollout reward remains
unchanged. For $M_g>0$ and $N>0$, we compute a separate entropic transport
plan $\Pi_g^\star$ on the real prediction--target IoU matrix $Q_g$, with
uniform row and column marginals $1/M_g$ and $1/N$, respectively. Instance
credits are computed as
$e_{g,i}=\sum_{j=1}^{N}\Pi^\star_{g,ij}Q_{g,ij}/
\sum_{j=1}^{N}\Pi^\star_{g,ij}$. Entropy regularization, solver settings,
branch-wise credit normalization, advantage modulation, and empty-set handling
remain unchanged.

\paragraph{Equivalent reward identity.}
For $M_g,N>0$, define the total real-match overlap mass as
\begin{equation}
S_g=
\sum_{i=1}^{M_g}\sum_{j=1}^{N}
P^\star_{g,ij}Q_{g,ij}.
\label{eq:transport-overlap-mass}
\end{equation}
When the marginal constraints in \eqref{eq:dect-marginals} are satisfied
exactly, every real prediction row and real target column has total mass
$1/(M_g+N)$. Substitution into \eqref{eq:transport-reward} gives
\begin{equation}
\mathrm{Prec}_g=\frac{M_g+N}{M_g}S_g,
\qquad
\mathrm{Rec}_g=\frac{M_g+N}{N}S_g,
\qquad
\boxed{R_g=2S_g}.
\label{eq:transport-reward-identity}
\end{equation}
The zero-reward convention following \eqref{eq:transport-reward} extends
this identity to $S_g=0$, including the case in which all real
prediction--target IoUs vanish. This equality provides an implementation
check rather than a new reward definition. Because finite Sinkhorn scaling
satisfies the marginals only up to the stopping tolerance, numerical checks
of the identity allow the corresponding residual error.

Introducing Lagrange multipliers $\boldsymbol\alpha$ and $\boldsymbol\beta$ for the row and column constraints gives
\begin{equation}
\mathcal L(P,\boldsymbol\alpha,\boldsymbol\beta)
=
\sum_{i,j}P_{ij}\overline Q_{g,ij}
-\varepsilon\sum_{i,j}P_{ij}\log P_{ij}
+\boldsymbol\alpha^\top(\mathbf a_g-P\mathbf 1)
+\boldsymbol\beta^\top(\mathbf b_g-P^\top\mathbf 1).
\label{eq:dect-lagrangian}
\end{equation}
Stationarity with respect to $P_{ij}$ yields
\begin{equation}
\overline Q_{g,ij}
-\varepsilon(\log P_{ij}+1)
-\alpha_i-\beta_j=0,
\end{equation}
and therefore
\begin{equation}
P_{ij}
=
u_i
\exp\!\left(\frac{\overline Q_{g,ij}}{\varepsilon}\right)
v_j,
\label{eq:dect-factorization}
\end{equation}
where the constant term is absorbed into the positive scaling factors $u_i$ and $v_j$. Defining the elementwise exponential kernel
\begin{equation}
\mathbf K_g=\exp\!\left(\frac{\overline Q_g}{\varepsilon}\right),
\end{equation}
the optimal plan takes the matrix form
\begin{equation}
P_g^\star
=
\operatorname{diag}(\mathbf u)\,
\mathbf K_g\,
\operatorname{diag}(\mathbf v).
\label{eq:dect-scaled-kernel}
\end{equation}

The marginal constraints determine the scaling vectors:
\begin{equation}
\mathbf u\odot(\mathbf K_g\mathbf v)=\mathbf a_g,
\qquad
\mathbf v\odot(\mathbf K_g^\top\mathbf u)=\mathbf b_g,
\end{equation}
which leads to the alternating Sinkhorn updates
\begin{equation}
\mathbf u\leftarrow
\mathbf a_g\oslash(\mathbf K_g\mathbf v),
\qquad
\mathbf v\leftarrow
\mathbf b_g\oslash(\mathbf K_g^\top\mathbf u),
\label{eq:dect-sinkhorn}
\end{equation}
where $\odot$ and $\oslash$ denote elementwise multiplication and division.
For numerical stability, the implementation maintains dual log-scalings
$\mathbf f=\varepsilon\log\mathbf u$ and
$\mathbf h=\varepsilon\log\mathbf v$ and alternates
\begin{equation}
\begin{aligned}
f_i&\leftarrow\varepsilon\left[
\log a_{g,i}-
\operatorname{LSE}_{j}
\left(\frac{\overline Q_{g,ij}+h_j}{\varepsilon}\right)
\right],\\
h_j&\leftarrow\varepsilon\left[
\log b_{g,j}-
\operatorname{LSE}_{i}
\left(\frac{\overline Q_{g,ij}+f_i}{\varepsilon}\right)
\right],\\
P_{g,ij}&=
\exp\left(\frac{\overline Q_{g,ij}+f_i+h_j}{\varepsilon}\right),
\end{aligned}
\label{eq:dect-log-sinkhorn}
\end{equation}
where $\operatorname{LSE}$ denotes log-sum-exp. This is algebraically
equivalent to \eqref{eq:dect-sinkhorn} but avoids explicitly forming large
entries of $\mathbf K_g$.

This factorization clarifies the role of entropy regularization in DECT. Without it, small IoU differences can induce nearly hard assignments, making instance credit sensitive to ambiguous or overlapping predictions. Entropy regularization instead produces the positive affinity kernel $\mathbf K_g$, allowing mass to be distributed among plausible matches before the marginal constraints reconcile competing assignments globally. Dustbin entries participate in the same normalization, so false-positive predictions and missed targets compete directly with real correspondences rather than being handled by separate heuristics.

Equation~\ref{eq:credit-map} converts the transport plan into instance-level credit. Real matches contribute in proportion to mask IoU, whereas dustbin assignments contribute no overlap credit. Thus, $P_g^\star$ resolves ambiguous correspondences globally while allowing unmatched predictions and targets to reduce the downstream reward and credit signals.

\paragraph{Representative transport and credit cases.}
The following examples use the fixed settings above. Matrices are rounded to
four decimals; their last row and last column denote the null prediction and
null target, respectively. We report the conditional instance credit
$e_i$ from \eqref{eq:credit-map}. For the normalized correction
$\widehat e_i$, we assume equal grounding/action output counts per predicted
instance; the final branch advantage is then
$A^b_{g,i}=A_g+0.5|A_g|\widehat e_i$.

\emph{Ambiguous matching.}
For two nearly interchangeable predictions and targets,
\begin{equation}
Q=
\begin{bmatrix}0.70&0.68\\0.68&0.70\end{bmatrix},\qquad
P^\star\approx
\begin{bmatrix}
0.1479&0.0991&0.0030\\
0.0991&0.1479&0.0030\\
0.0030&0.0030&0.4940
\end{bmatrix}.
\label{eq:dect-example-ambiguous}
\end{equation}
The off-diagonal mass preserves matching uncertainty. Both predictions receive
the same credit, $e=(0.6837,0.6837)$, so $\widehat e=(0,0)$. This avoids an
arbitrary instance-level preference, while the rollout reward is
$R_g=0.6837$.

\emph{Duplicate predictions.}
For two predictions competing for one target,
\begin{equation}
Q=
\begin{bmatrix}0.88\\0.62\end{bmatrix},\qquad
P^\star\approx
\begin{bmatrix}
0.3103&0.0231\\
0.0230&0.3103\\
0.0000&0.3333
\end{bmatrix}.
\label{eq:dect-example-duplicate}
\end{equation}
Most target mass is assigned to the stronger prediction, whereas the weaker
duplicate is sent to the null target. The credits are
$e=(0.8191,0.0429)$ and $\widehat e=(1,-1)$, so the former receives
$A_g+0.5|A_g|$ and the latter $A_g-0.5|A_g|$; the set reward is
$R_g=0.5747$.

\emph{Missed target.}
For one prediction and two targets,
\begin{equation}
Q=
\begin{bmatrix}0.84&0.12\end{bmatrix},\qquad
P^\star\approx
\begin{bmatrix}
0.3324&0.0001&0.0009\\
0.0009&0.3333&0.3325
\end{bmatrix}.
\label{eq:dect-example-missed}
\end{equation}
The second target is explained almost entirely by the null-prediction row.
The existing prediction has $e_1=0.8376$ and, because it is the only
prediction, $\widehat e_1=0$. The missed target instead lowers recall and
the shared rollout reward to $R_g=0.5584$, thereby affecting all sampled
outputs through $A_g$ even though no instance-specific span exists for the
missing object.

\subsection{DECT-GRPO Optimization Details}
\label{sec:rl-protocol}

\paragraph{Credit-map normalization.}
For rollout $g$, the instance credit $e_{g,i}$ in \eqref{eq:credit-map} is broadcast to the grounding-token span $S_{g,i}$ and valid action indices $\mathcal D_{g,i}$. Because different instances can contribute different numbers of grounding-token and action-likelihood terms, the two branches normalize the credits separately. Let
\begin{equation}
w^y_{g,i}=|S_{g,i}|,
\qquad
w^a_{g,i}=|\mathcal D_{g,i}|,
\qquad
\overline e_g^b=
\frac{\sum_i w^b_{g,i}e_{g,i}}
{\sum_i w^b_{g,i}},
\end{equation}
for $b\in\{y,a\}$. The normalized instance credit is
\begin{equation}
\widehat e^b_{g,i}
=
\frac{e_{g,i}-\overline e_g^b}
{\max_j|e_{g,j}-\overline e_g^b|},
\label{eq:credit-normalization}
\end{equation}
when the denominator exceeds $10^{-6}$; otherwise, the correction is set to zero. The credit maps are constructed by assigning
$\mathbf c_g^y[S_{g,i}]=\widehat e^y_{g,i}$ and
$\mathbf c_g^a[\mathcal D_{g,i}]=\widehat e^a_{g,i}$.
This output-count-weighted centering preserves the branch-wise mean advantage after instance-level modulation, as shown in \eqref{eq:advantage-conservation}. It requires valid, nonoverlapping grounding spans and valid-point masks for the action branch.

\paragraph{Sign-preserving modulation.}
For $A_g\neq 0$, the instance-modulated advantage satisfies
\[
\frac{A^b_{g,\nu}}{A_g}
=
1+\lambda_c\,\operatorname{sgn}(A_g)\mathbf c_g^b[\nu]
\in [1-\lambda_c,\,1+\lambda_c].
\]
Thus, for $0\leq\lambda_c<1$, the correction preserves the sign of the
shared advantage. With the default $\lambda_c=0.5$, the ratio lies in
$[0.5,1.5]$. When $A_g=0$, the correction also vanishes. The mechanism
therefore differentiates the strength of shared rollout feedback across
instances without assigning independent advantage signs.

\paragraph{Empty outputs and set rewards.}
For positive-target inputs with no predicted objects, the rollout reward is set to
\begin{equation}
R_g=0,\qquad M_g=0,\;N>0.
\end{equation}
For no-target inputs, the reward depends only on the predicted object count,
\begin{equation}
R_g=\frac{1}{1+M_g},\qquad N=0.
\end{equation}
For these no-target examples, grounding advantages are multiplied by the configured weight $\alpha_{\mathrm{neg}}$. The action branch is disabled because the reward evaluates object count rather than contour quality, even when the rollout contains predicted objects. Missed targets have no corresponding grounding span or action trajectory and therefore influence optimization through target coverage, the set reward $R_g$, and the shared group-relative advantage $A_g$.

\paragraph{Likelihood ratios and clipped objectives.}
During rollout collection, we store grounding and action log-probabilities under the behavior policy. During replay, the current log-probabilities are recomputed on the same sampled outputs. Their likelihood ratios are
\begin{equation}
\rho^y_{g,n}
=
\exp\!\left(
\ell^y_{g,n}-\ell^{y,\mathrm{old}}_{g,n}
\right),
\qquad
\rho^a_{g,\xi}
=
\exp\!\left(
\ell^a_{g,\xi}-\ell^{a,\mathrm{old}}_{g,\xi}
\right).
\label{eq:policy-ratios}
\end{equation}
Both branches use the clipped surrogate from \eqref{eq:joint-loss}, defined explicitly as
\begin{equation}
\mathcal S_{\mathrm{clip}}(\rho,A)
=
\min\!\left(
\rho A,\,
\operatorname{clip}(\rho,1-c,1+c)A
\right).
\label{eq:clipped-surrogate}
\end{equation}

For grounding, the instance-modulated token advantages from \eqref{eq:instance-advantage} yield
\begin{equation}
\mathcal L_g^y
=
-\frac{1}{L_g}
\sum_{n=1}^{L_g}
\mathcal S_{\mathrm{clip}}(\rho^y_{g,n},A^y_{g,n})
+
\beta_{\mathrm{KL}}\mathcal K_g.
\label{eq:grounding-policy-loss}
\end{equation}
For geometric actions,
\begin{equation}
\mathcal L_g^a
=
-\frac{1}{|\mathcal D_g|}
\sum_{\xi\in\mathcal D_g}
\mathcal S_{\mathrm{clip}}(\rho^a_{g,\xi},A^a_{g,\xi}),
\label{eq:action-policy-loss}
\end{equation}
where $\mathcal D_g$ contains all valid two-dimensional actions in the rollout. Padding points do not contribute to the action likelihood or loss.

\paragraph{Grounding reference regularization.}
The grounding branch is regularized toward a reference policy. Let
\begin{equation}
d_{g,n}
=
\operatorname{sg}\!\left(
\ell^{y,\mathrm{ref}}_{g,n}
\right)
-
\ell^y_{g,n}.
\end{equation}
We use
\begin{equation}
\mathcal K_g
=
\frac{1}{L_g}
\sum_{n=1}^{L_g}
\left(
\exp(d_{g,n})-d_{g,n}-1
\right).
\label{eq:grounding-reference}
\end{equation}
This nonnegative exponential surrogate is applied only to grounding; the continuous action branch uses no reference-policy regularizer.

\paragraph{Fixed-sample replay.}
Rollout collection stores the sampled grounding output, pre-$\tanh$ action latents, behavior-policy log-probabilities, valid-point masks, boxes, and historical contour states. During replay, these sampled quantities remain fixed while the current VLIM, EASS, and GAD features and action means are recomputed with gradients enabled. Rewards, rasterized masks, transport plans, and credit maps are detached from gradient computation.

Consequently, the action branch is optimized through score-function gradients of the sampled continuous actions rather than differentiation through contour rasterization, mask rewards, or transport matching. Gradients from the action likelihood propagate through GAD and its trainable observation pathways, including EASS and the trainable VLIM features.

\paragraph{Update aggregation.}
Rollouts with zero group-relative advantage or without scoreable grounding outputs are omitted from the policy loss, including the grounding reference term. The action loss is additionally omitted when $M_g=0$, $N=0$, or $|\mathcal D_g|=0$. Each remaining rollout retains its $1/G$ weight in \eqref{eq:joint-loss}; grounding and action losses are normalized independently before applying $\lambda_a$. Across gradient-accumulation steps, gradients are averaged over the accumulation count and across devices, then clipped before the optimizer update. Each sampled group is used for a single policy update.

\subsection{Additional Ablation Studies}
\label{app:additional_ablations}

\subsubsection{State Conditioning in Semantic Routing}
\label{app:state_conditioning}

\paragraph{Experimental setup.}
We compare four semantic-routing variants to distinguish learned layer
selection from conditioning on static instance attributes and evolving
contour geometry. All variants retain Spatial Visual Sampling (SVS),
contour-aligned spatial re-observation, and the default action horizon
$K=2$ and $T=4$. They follow the same Stage-II SFT protocol from the same
Stage-I initialization and are evaluated with identical grounding outputs
and initial contours, as in Table~\ref{tab:eass_ablation}. No RL post-training
is used in this comparison.

The \emph{predefined schedule} retains SVS and spatial re-observation but uses
the predefined VLIM layer schedule instead of learned semantic routing.
\emph{Slot-conditioned routing} retains evolution--depth slot embeddings and
the normalized evolution index, but masks instance-specific and dynamic
geometric descriptors at the query input.
\emph{Static-instance-conditioned routing} additionally retains log box
aspect ratio, log box area, and normalized point allocation. \emph{Full
state-conditioned routing} further includes contour irregularity and the mean
and maximum displacement from the preceding evolution iteration, recovering
Full EASS. These descriptors follow Appendix~\ref{sec:implementation-details}.

For the three learned-routing variants, masked query dimensions are set to
zero during both training and inference without changing the query-network
width. Semantic-key construction, candidate layers, straight-through
selection, the routing-prior schedule, and routing regularizers remain
unchanged. Only routing-query inputs are ablated; geometry in SVS, contour
tokens, and instance context is retained. Because semantic keys depend on
VLIM features, slot-conditioned routing can still vary across
image--expression inputs rather than imposing a dataset-wide fixed schedule.

\paragraph{Evaluation.}
We report gIoU and cIoU on gRefCOCO val under the common-grounding protocol of
Table~\ref{tab:eass_ablation}. We additionally report mean Boundary IoU on the
fixed positive-target expression subset used in
Figure~\ref{fig:contour_evolution}, denoted $\mathrm{bIoU}_{+}$. It is
evaluated after eight action steps using expression-level union masks,
original image resolution, and the boundary-width parameter of $2\%$ of the
image diagonal specified in Appendix~\ref{app:training_eval}. Empty
predictions receive zero and remain in the average.

\begin{table}[H]
\centering
\caption{\textbf{State conditioning in semantic routing on gRefCOCO val
(\%).} All variants retain SVS and spatial re-observation. gIoU and cIoU use
the validation split; $\mathrm{bIoU}_{+}$ uses the fixed positive-target
subset from Figure~\ref{fig:contour_evolution} at action step~8. The Full EASS
boundary score is the Figure~\ref{fig:contour_evolution} endpoint ($65.98\%$),
rounded to one decimal place. Bold denotes the best score in each column.}
\label{tab:state_conditioning}
\small
\setlength{\tabcolsep}{5pt}
\renewcommand{\arraystretch}{1.10}
\begin{tabular}{lccc}
\toprule
Routing variant
& gIoU $\uparrow$
& cIoU $\uparrow$
& $\mathrm{bIoU}_{+}$ $\uparrow$ \\
\midrule
Predefined schedule
& 74.2
& 63.5
& 61.4 \\
Slot-conditioned
& 74.7
& 64.2
& 62.2 \\
Static-instance-conditioned
& 76.5
& 65.1
& 64.3 \\
Full state-conditioned
& \textbf{77.6}
& \textbf{67.4}
& \textbf{66.0} \\
\bottomrule
\end{tabular}
\end{table}

\paragraph{Analysis.}
Table~\ref{tab:state_conditioning} shows modest gains from slot-conditioned
routing over the predefined schedule, followed by further improvements when
static instance attributes are included. Full state conditioning adds 1.1
gIoU, 2.3 cIoU, and 1.7 $\mathrm{bIoU}_{+}$ points over the static variant.
With SVS and spatial re-observation retained, this comparison supports the
value of evolving contour descriptors for semantic routing beyond
evolution--depth slots and static instance attributes. The boundary
improvement on the fixed positive-target subset is consistent with more
effective semantic conditioning during contour refinement.

\subsubsection{Grounding and Contour-Action Optimization}
\label{app:pg_objectives}

\paragraph{Experimental setup.}
We ablate the two policy-gradient (PG) terms while retaining the complete
architecture. Let $\mathcal L_{\mathrm{PG}}^{y}$ and
$\mathcal L_{\mathrm{PG}}^{a}$ denote the aggregated grounding-token and
contour-action clipped-surrogate losses, respectively, excluding reference
regularization and the action coefficient. We use
\begin{equation}
\mathcal L_{\eta_y,\eta_a}
=
\eta_y\mathcal L_{\mathrm{PG}}^{y}
+
\eta_a\lambda_a\mathcal L_{\mathrm{PG}}^{a}
+
\beta_{\mathrm{KL}}\mathcal K.
\label{eq:pg_objective_ablation}
\end{equation}
The switches $(\eta_y,\eta_a)=(1,0)$, $(0,1)$, and $(1,1)$ give
\emph{grounding-PG only}, \emph{action-PG only}, and \emph{joint DECT-GRPO},
respectively. $\mathcal K$ retains the original grounding reference penalty
and aggregation rules in Appendix~\ref{sec:rl-protocol} for all RL variants.
SFT-only denotes the checkpoint before post-training.

All RL variants use the same SFT initialization, training data, rollout
settings, transport reward, dustbin treatment, instance credits, and
optimization budget. Both grounding and actions are sampled regardless of
which PG term is retained. Loss coefficients and validity masks are unchanged,
and no SFT losses are added. No-target rollouts receive no action PG loss. This
is an objective ablation, not module freezing: shared VLIM features can change
even when one PG term is absent.

\paragraph{End-to-end and fixed-grounding evaluation.}
Table~\ref{tab:pg_objectives}(a) evaluates the full gRefCOCO validation split
using each model's own grounding. Panel (b) fixes the same complete SFT
grounding sequences, boxes, and initial contours for every model on the fixed
positive-target subset used in Figure~\ref{fig:contour_evolution}. Models
recompute their own VLIM features, EASS observations, and eight-step contour
actions without regenerating grounding outputs. Thus, upstream predictions
are controlled, but learned observation pathways are not frozen.

On this subset, $\mathrm{mIoU}_{+}$ averages expression-level union-mask IoU,
while $\mathrm{bIoU}_{+}$ follows the boundary protocol above. Empty
predictions remain in both averages with zero scores; no model-specific
filtering is applied. The SFT-only boundary reference is shared with Full EASS
in Table~\ref{tab:state_conditioning}. Panel (b) is a separate controlled
evaluation, not a positive-target breakdown of panel (a).

\begin{table}[H]
\centering
\caption{\textbf{Ablation of grounding and contour-action PG objectives
(\%).} (a) End-to-end evaluation on the full gRefCOCO validation split using
each model's own grounding. (b) Eight-step contour evaluation on the fixed
positive-target subset from Figure~\ref{fig:contour_evolution}, with the same
SFT grounding and initial contours for every model. Bold denotes the best score
in each column.}
\label{tab:pg_objectives}
\small
\setlength{\tabcolsep}{4pt}
\renewcommand{\arraystretch}{1.10}
\begin{tabular}{lrrrrrr}
\toprule
& \multicolumn{4}{c}{\textbf{(a) End-to-end}}
& \multicolumn{2}{c}{\textbf{(b) Fixed grounding}} \\
\cmidrule(lr){2-5}\cmidrule(lr){6-7}
Variant
& gIoU $\uparrow$
& cIoU $\uparrow$
& N-acc $\uparrow$
& F1@0.5 $\uparrow$
& $\mathrm{mIoU}_{+}$ $\uparrow$
& $\mathrm{bIoU}_{+}$ $\uparrow$ \\
\midrule
SFT only
& 77.6
& 67.4
& 74.6
& 73.8
& 78.4
& 66.0 \\
Grounding-PG only
& 81.4
& 70.2
& 80.7
& 79.8
& 79.0
& 66.7 \\
Action-PG only
& 79.8
& 69.5
& 76.3
& 76.1
& 80.5
& 68.5 \\
Joint DECT-GRPO
& \textbf{83.5}
& \textbf{72.3}
& \textbf{83.2}
& \textbf{82.4}
& \textbf{81.2}
& \textbf{69.3} \\
\bottomrule
\end{tabular}
\end{table}

\paragraph{Analysis.}
Both single-PG objectives improve over SFT, but their relative performance
differs between the two evaluation settings. Grounding-PG-only training
achieves higher end-to-end scores, whereas action-PG-only training yields
better region overlap and boundary quality under fixed SFT grounding. Joint
DECT-GRPO achieves the highest scores in both panels. Relative to grounding-PG
only, it improves end-to-end gIoU and F1@0.5 by 2.1 and 2.6 points,
respectively. Under fixed grounding, it retains gains of 2.2
$\mathrm{mIoU}_{+}$ and 2.6 $\mathrm{bIoU}_{+}$ points, supporting a
refinement benefit from including the action objective beyond changes in
predicted boxes and referent counts at evaluation. Joint training also
improves over action-PG only under both protocols, supporting the retention of
both PG terms within the evaluated training budget. These comparisons support
the two training objectives without attributing their gains to isolated
modules.

\FloatBarrier

\subsection{Limitations and Scope}
\label{sec:limitations}

\paragraph{Scope of the closed loop.}
In the current formulation, ``closed-loop VLA'' specifically denotes
feedback between the evolving geometry of initialized contours and
state-conditioned visual--semantic readouts. VLIM determines the instance
set and initial boxes before contour evolution; the subsequent action space
refines these instances but does not regenerate boxes or introduce
instance-birth and instance-deletion actions within the same inference
rollout. Consequently, an initially missed referent---including the
$M=0$ case---cannot be recovered through later contour updates. DECT-GRPO
improves the grounding policy across training rollouts, but is not intended
to imply online bidirectional revision of grounding and segmentation.
This scoped design isolates geometric correction and keeps inference
stable; extending the action space with re-grounding and cardinality-changing
operations is a natural direction for a broader closed-loop policy.

\paragraph{Contour representation.}
The fixed ordered-vertex state provides efficient, interpretable boundary
updates, but retains the initialized contour topology and point budget.
Highly disconnected regions, holes, or exceptionally thin structures may
therefore benefit from topology-changing actions or adaptive vertex
insertion. These extensions are complementary to EASS and DECT-GRPO rather
than changes to their underlying perception and credit-assignment principles.

\end{document}